\documentclass[default,iicol]{sn-jnl}

\pdfoutput=1 
\usepackage{booktabs}

\usepackage{graphicx}%

\usepackage{amsmath,amssymb,amsfonts}%
\usepackage{makecell}
\usepackage[dvipsnames]{xcolor}
\usepackage{float}

\usepackage[T1]{fontenc}
\usepackage{lmodern}
\usepackage{microtype}

\usepackage{amsthm}%
\usepackage{mathrsfs}%
\usepackage[title]{appendix}%
\usepackage{xcolor}%
\usepackage{textcomp}%
\usepackage{manyfoot}%
\usepackage{booktabs}%

\usepackage{algorithm}
\usepackage{algorithmic}
\usepackage{listings}%
\usepackage{subfigure}
\usepackage{enumerate}
\usepackage{textcomp}

\usepackage{subfigure}
\usepackage{colortbl}
\usepackage{array}
\usepackage{xcolor}

\usepackage{multirow}

\usepackage{arydshln}
\usepackage{multicol}

\usepackage{color}
\usepackage{pifont}
\newcommand{\cmark}{\ding{51}}
\newcommand{\xmark}{\ding{55}}

\usepackage{cite}

\usepackage{geometry}
\usepackage{colortbl}

\usepackage[numbers]{natbib}
\begin{document}

\title[Article Title]{ProSGNeRF: Progressive Dynamic Neural Scene Graph with Frequency Modulated Foundation Model in Urban Scenes}


\author[1]{\fnm{Tianchen} \sur{Deng}}
\author[1]{\fnm{Yanbo} \sur{Wang}}
\author[1]{\fnm{Yejia} \sur{Liu}}
\author[1]{\fnm{Chenpeng} \sur{Su}}
\author[1]{\fnm{Jingchuan} \sur{Wang}}
\author[1]{\fnm{Hesheng} \sur{Wang}}
\author[2]{\fnm{Danwei} \sur{Wang}}
\author[3]{\fnm{Shao-Yuan} \sur{Lo}}
\author[1]{\fnm{Weidong} \sur{Chen}*}
\affil[1]{\orgname{Shanghai Jiao Tong University}}
\affil[2]{\orgname{Nanyang Technological University}}
\affil[3]{\orgname{National Taiwan University}}


\vspace{10cm}

\abstract{
Implicit neural representation has demonstrated promising results in 3D reconstruction in various scenes. However, existing approaches either struggle to model fast-moving objects or are incapable of handling large-scale camera ego-motion in urban environments. This leads to low-quality synthesized views of the large-scale urban scenes. In this paper, we aim to jointly solve the problems caused by large-scale scenes and fast-moving vehicles, which are more practical and challenging. To this end, we propose a progressive scene graph network architecture to learn the local scene representations of dynamic objects and global urban scenes. The progressive learning architecture dynamically allocates a new local scene graph trained on frames within a temporal window, with the window size automatically determined, allowing us to scale up the representation to large-scale scenes. Besides, according to our observations, fast-moving objects are observed only in a few frames, which leads to a significant decline in reconstruction accuracy for dynamic objects. Therfore, We introduce a foundation-guided object representation that extracts object-centric visual priors and conditions the density and color decoders in normalized object coordinates. We further propose a frequency-progressive regularization strategy that gradually exposes high-frequency positional, directional, and pose encodings during training, reducing overfitting to sparse object observations. Experimental results demonstrate that our method achieves state-of-the-art view synthesis accuracy, object manipulation, and scene roaming ability in various scenes. The code will be open-sourced on \href{https://github.com/dtc111111/prosgnerf}{https://github.com/dtc111111/prosgnerf}.
}


\maketitle


\newpage


\begin{figure*}[t]
  \centering
   \includegraphics[width=\linewidth]{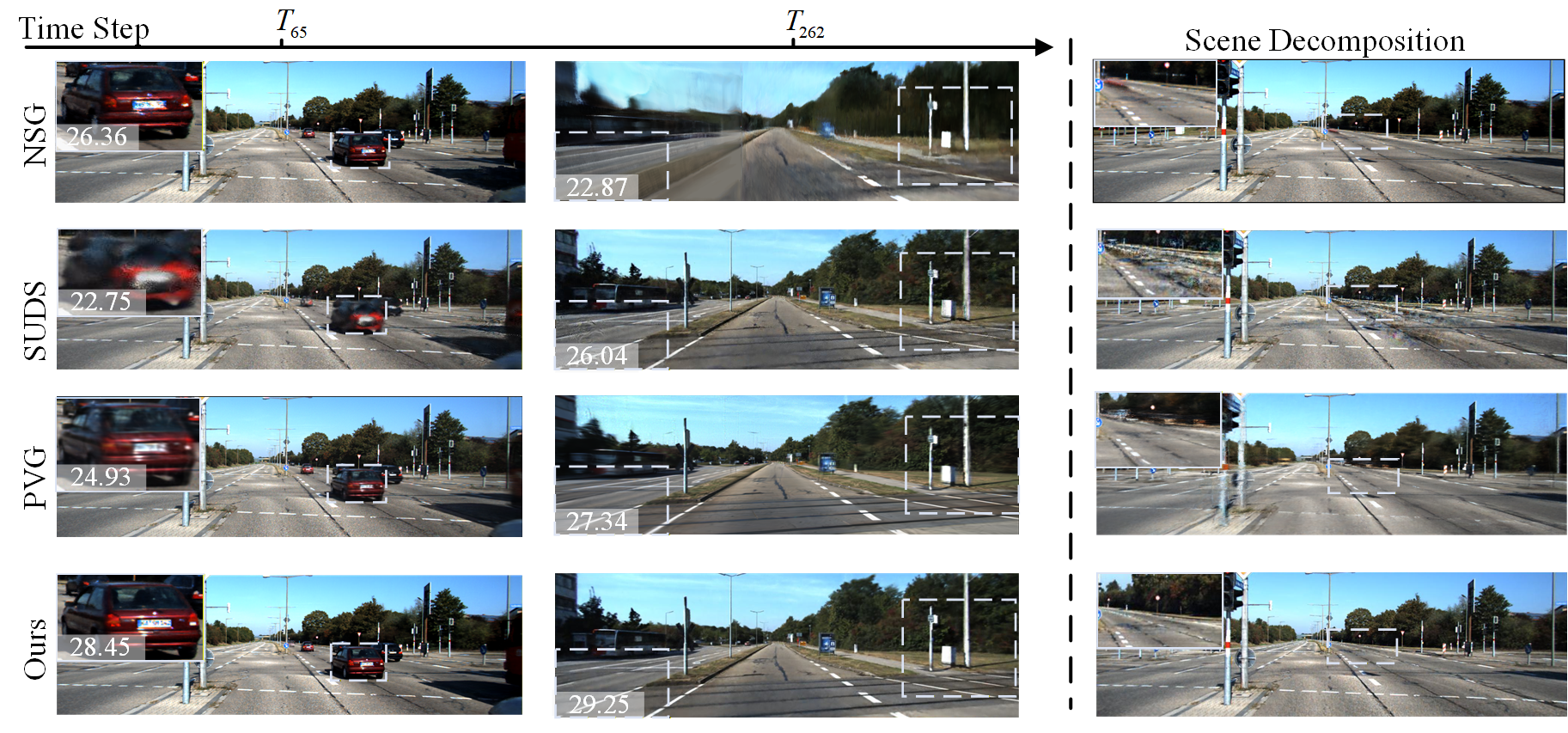}
    \caption{Urban scene reconstruction and editing with ProSGNeRF. We show our view synthesis in different time steps (65,262) and scene decomposition results. Our approach significantly improves the view synthesis performance in real-world urban scenes containing multiple dynamic objects and large-scale camera ego-motion. We highlight and enlarge objects in the first column of images, providing the corresponding object PSNR in the top left corner. Scene PSNR is provided in the second column.
    }
   \label{fig:progressive_1}
\end{figure*}

\section{Introduction}
\label{sec:introduction}

Urban scene reconstruction and novel view synthesis are fundamental tasks for autonomous driving~\cite{Deng_2026_CVPR}, robotic navigation~\cite{deng2025best3dscenerepresentation}, city-scale simulation, and virtual/augmented reality~\cite{fmgs,cen2025segment}. In autonomous driving, photo-realistic and controllable reconstruction of road scenes can support closed-loop simulation, corner-case generation, and safety validation at a lower cost than large-scale real-world testing.

NeRF~\cite{NeRF} has shown promising results for view synthesis on static and object-centric scenes. Some recent works improve the original NeRF and achieve large-scale urban scene reconstruction. Block-nerf~\cite{blocknerf} pre-divides the scene into multiple blocks with per-block update for virtual drive-through reconstruction. Mega-nerf \cite{meganerf} uses expanded octree architecture for large-scale photo-realistic fly-through scenarios. BungeeNeRF~\cite{bungeenerf} designs a growing model with residual block structure for city-level scene reconstruction. \cite{Localrf,mneslam} focus on incremental scene representation, where multiple MLP modules are employed to model the entire scene, and the representations are fused through an online distillation strategy. NSG~\cite{nsg} introduces a scene graph architecture to represent the scene, while SUDS~\cite{suds} factorizes the urban scene into three separate hash table data structures to efficiently encode static, dynamic, and far-field radiance fields. Mars~\cite{mars} proposes a modular framework with instance-aware mask and  separately with independent networks so that the static model the dynamic properties of instances and background. EmerNeRF~\cite{emernerf} represents the scenes into static and dynamic fields. It also uses the flow field to aggregate multi-frame features, improving the rendering precision of dynamic objects. 
Although existing methods have made improvements, they are still insufficient for real-world urban scenes. We identify three key challenges in dynamic urban view synthesis. First, urban scenes contain multiple dynamic objects with independent motions. Modeling these objects as transient changes in a global radiance field often leads to blurred object boundaries and ghosting artifacts. Second, long driving sequences with large ego-motion require a representation whose capacity can grow with the explored space. A single global field with fixed capacity may suffer from reduced fidelity or excessive memory consumption as the scene scale increases. Third, each dynamic object is usually observed from only a small number of frames due to fast relative motion and occlusion. This sparse-view setting makes object-centric geometry and appearance reconstruction highly under-constrained.

These observations motivate us to revisit dynamic urban view synthesis from two coupled perspectives. First, long driving sequences should not be represented by a single global radiance field with fixed capacity. The representation needs to grow progressively along the camera trajectory while preserving consistency across neighboring regions. Second, dynamic objects should not be treated merely as transient appearance changes in the global field. Since each moving instance is often observed from only a few viewpoints, object-centric priors and sparse-view regularization are necessary for reconstructing high-quality dynamic objects.

To this end, we propose ProSGNeRF, a progressive dynamic neural scene graph for urban view synthesis. ProSGNeRF allocates local scene graphs along the camera trajectory and decomposes each graph into background, dynamic object, and far-field components.  In each local scene graph, we incorporate multiresolution hash encoding into the background representation to better model low-level architectural details. Dynamic objects are represented in normalized object coordinates and conditioned on foundation-model visual priors extracted from their training observations.  We ultilize the visual foundation model DINO V2~\cite{dino} to extract appearance and shape codes for dynamic objects with the prior information to enhance the reconstruction accuracy of dynamic objects. A frequency-progressive regularization strategy is further introduced to stabilize sparse-view object optimization by gradually increasing the visible frequency bands of positional, directional, and pose encodings.
The Segment Anything Model~\cite{sam} has demonstrated its remarkable ability to revolutionize 2D segmentation. We leverage the promptable segmentation method to predict masks for dynamic objects and far-field scenarios, which can reduce the need for human annotation. We also use lidar projection supervision for geometry consistency. To validate our proposed framework, we evaluate the effectiveness of the method on different outdoor urban datasets, such as KITTI~\cite{kitti}, VKITTI~\cite{vkitti}, Waymo~\cite{waymo}, nuScenes~\cite{nuScenes} datasets, and do exhaustive evaluations and ablation experiments on these datasets. Our system demonstrates superior performance compared to existing methods that scene reconstruction on urban scenes.
In summary, our contributions are shown as follows:
\begin{itemize}
    \item We propose a novel editable neural scene reconstruction method, ProSGNeRF, which decomposes dynamic, background, and far-field into a progressively neural scene graph with decoupled object transformations and scene representations and also supports object manipulation and scene roaming.
    
    \item We propose a progressive neural scene graph representation architecture for the large-scale scenes, We dynamically instantiate local neural scene graphs with temporal sliding windows, enabling scalability to handle arbitrary large-scale scenes and long sequences.
    \item In each neural scene graph, the scene is decomposed into multiple dynamic objects and a background component. For background representation, we adopt a multiresolution hash encoding strategy to effectively capture information at different frequency levels, enabling accurate modeling of both geometric structure and color appearance at fine spatial scales.
    \item Due to the sparse observation of objects in the urban scene, we design a frequency auto-encoder network with vision foundation model to efficiently encode shape and appearance features for the objects of interest and perform frequency modulation for sparse view inputs. In addition, to better train our vision foundation model, we construct a curated dataset consisting of hundreds of thousands of 2D images and corresponding 3D spatial information from the KITTI and Waymo datasets.
    
\end{itemize}
\begin{figure*}[t]
  \centering
   \includegraphics[width=\linewidth]{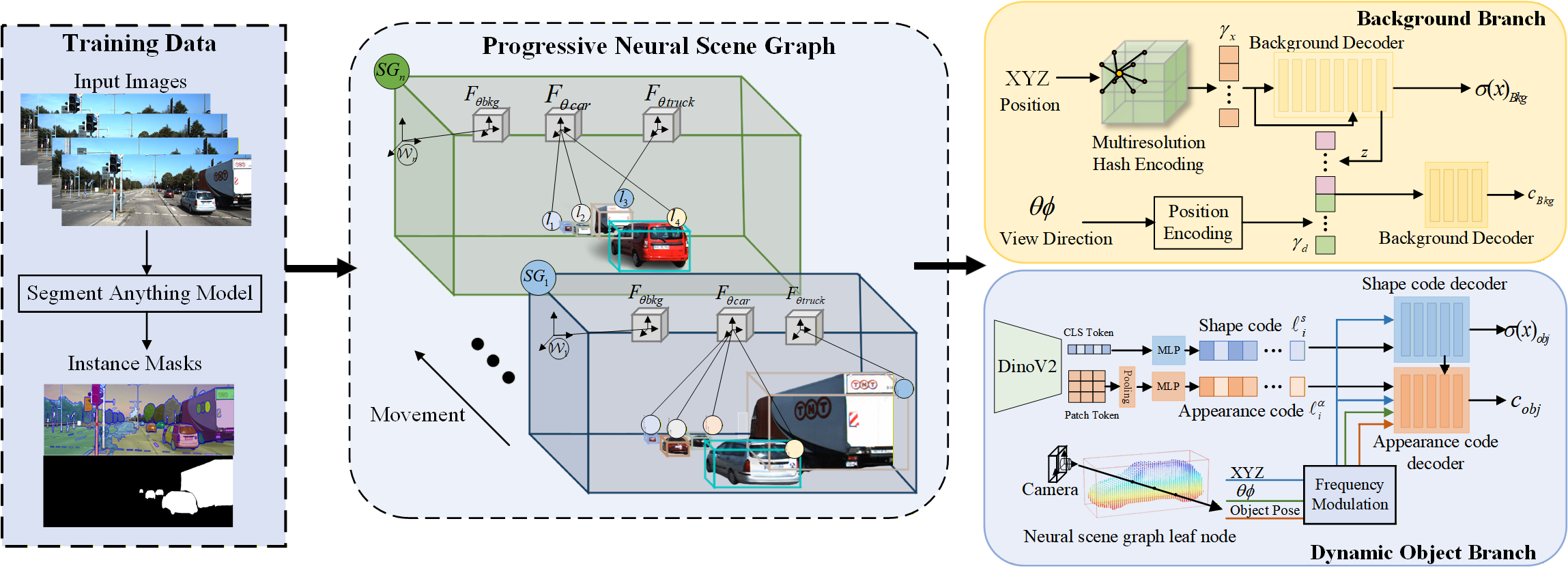}
    \caption{The isometric view of the proposed method, ProSGNeRF. We employed a 2D segmentation network, SAM, to preprocess the training data and generate accurate masks for dynamic objects. We propose a progressive neural scene graph architecture that dynamically allocates local neural scene graph (box). The entire scene is decomposed into three parts: background, dynamic objects, and far-field. We design separate networks for background and objects and introduce a far-field loss for regularization. Nodes $l_i$ represent individual dynamic objects. $F_{bkg}$ models the static background scene and $F_{obj}$ models movable foreground objects in local object-centric coordinate frames.}
   \label{fig: pipeline}
\end{figure*}
\section{Related Work}
Recently, with the introduction and advancement of Neural Radiance Fields \cite{NeRF}, a broad array of subsequent research has been inspired, building upon the original approach. We review the related work in the areas of implicit scene representations, dynamics, and large-scale scene representation. 

\subsection{Neural Scene Representations}

Neural Radiance Fields (NeRF) introduce an implicit neural representation that maps 3D positions and viewing directions to density and color, enabling photo-realistic novel view synthesis from posed images.
NeRF++~\cite{nerf++}, Mip-nerf~\cite{mipnerf}, and Mip-NeRF 360~\cite{mipnerf360} further improve the representation ability for unbounded environments. Those methods use the non-linear scene parameterization method to model unbounded scenes. \cite{srf} proposes a spherical radiance field (SRF) for efficient novel view synthesis in unbounded scenes. NeRF-W \cite{nerfw} incorporate per-image embeddings to model appearance variation of the outdoor scenes. NeRF-Der++~\cite{nerfdet} and Surface-SOS~\cite{surfacesos} combine implicit scene representation with object detection method for semantic and geometry consistency. To further accelerate the rendering process, 3DGS~\cite{3dgs} proposes to represent the entire scene using Gaussian primitives.

\subsection{Dynamic and Object-aware Neural Rendering}

There are some methods that focus on dynamic object reconstruction, such as Neural 3D Video Synthesis~\cite{n3dvs} and Space-time Neural radiance Fields~\cite{spacetime}. These approaches encode time information into scene representation. \cite{nsff} and \cite{dynerf} incorporate 2D optical flow with warping-based regularization losses
to enforce multi-view consistency and dynamic object perception. Nerfies~\cite{nerfies} and Hypernerf~\cite{hypernerf} design a per-frame deformation field for dynamic humans. These methods are limited to single-object scenes, whereas our method scales beyond synthetic and indoor scenes to urban scenes.

\subsection{Large-scale Urban Neural Rendering}

For large-scale scene representation, \cite{bungeenerf} proposes Bungeenerf with a progressively growing model for city-level rendering. Block-NeRF~\cite{blocknerf} and Mega-NeRF~\cite{meganerf} decompose the scene into multiple local radiance fields for large-scale scenes. However, these methods can only deal with static environments. Dynamic objects are prevalent in real-world urban road scenes. Panoptic-NeRF~\cite{panopticnerf} and Panoptic Neural Fields~\cite{pnf} produce panoptic segmentation jointly with scene representations, achieving urban scene reconstruction.\cite{plgslam} proposes incremental scene representation method for large-scale indoor environments.

NSG~\cite{nsg}, SUDS~\cite{suds}, and MARS~\cite{mars}  are the most relevant works to ours. NSG adopts neural scene graphs and dynamic object bounding boxes to decouple the scene but cannot handle ego-motion. Besides, NSG needs over 1TB of memory to represent a 30-second video. SUDS~\cite{suds} uses  Foundation model~\cite{dinov2} to distill scene features and obtain semantic information and scene flow. Their methods can handle camera ego-motion and static objects but struggle with high-dynamic objects. With the proposal of 3D gaussian splating~\cite{3dgs} Some 3DGS-based methods, such as DrivingGaussian~\cite{drivinggaussian},  StreetGaussian~\cite{streetgaussian} and~\cite{vpgsslam,pvg}, also focus on urban scene reconstruction. Nevertheless, long-sequence object-aware reconstruction remains challenging due to the increasing number of primitives, dynamic-object association, and memory management over extended trajectories. Our approach effectively addresses large-scale camera ego-motion and dynamic objects and long video sequences, enabling urban scene editing useful for constructing simulators for autonomous driving.

\section{Method}

The overall pipeline of ProSGNeRF is illustrated in ~\ref{fig: pipeline}. Given a driving sequence, the inputs to our method include sequential RGB images $\{I_i\}_{i=1}^{M}$, camera intrinsics $\{K_i\}_{i=1}^{M}$, camera poses $\{\mathbf{T}^{w\leftarrow c}_i\}_{i=1}^{M}$, sparse LiDAR point clouds $\{\mathcal{P}_i\}_{i=1}^{M}$, instance-level 3D bounding boxes $\{\mathcal{B}_{i,k}\}$, and 2D masks $\{\mathcal{M}_{i,k}\}$ for dynamic objects and far-field regions. The 3D boxes are obtained from dataset annotations and off-the-shelf 3D detector, while the 2D masks are generated by a promptable segmentation model using projected boxes or region prompts.

ProSGNeRF represents a long driving sequence as a set of overlapping local neural scene graphs. Each local graph decomposes the scene into three components: a static background branch, a set of dynamic object nodes, and a far-field branch. The background branch models static scene content with a neural radiance field, while each dynamic object node is represented in a normalized object-centric coordinate system and conditioned on foundation-model visual priors extracted from its training observations. During rendering, rays are evaluated through the corresponding local graphs, and overlapping graph outputs are fused to maintain consistency across neighboring regions.

\begin{figure*}[t]
  \centering
   \includegraphics[width=\linewidth]{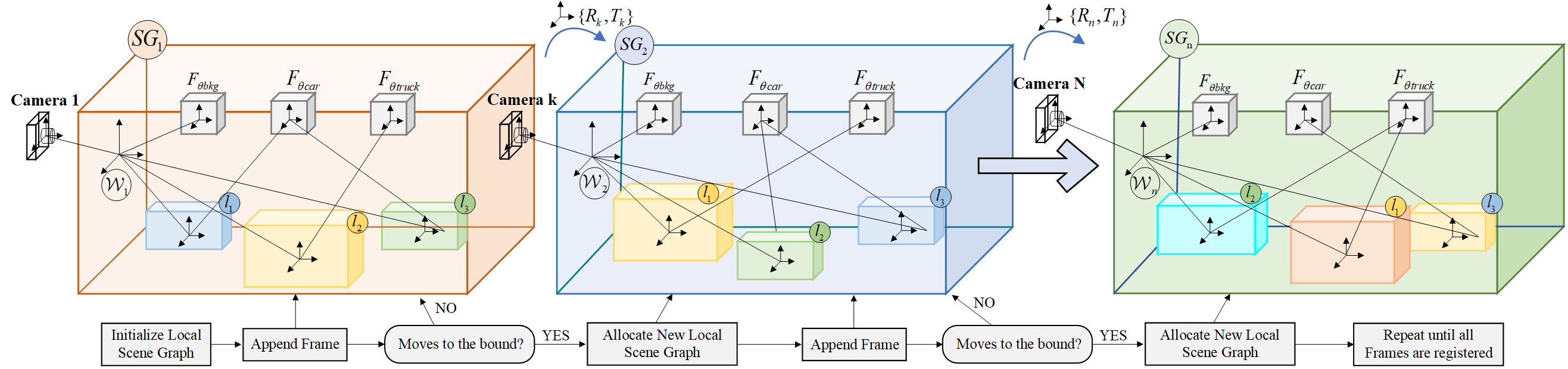}
    \caption{The isometric view of the proposed progressive neural scene graph. This scene representation uses a progressive scheme that dynamically allocates a local neural scene graph (box) based on the camera pose. Adjacent local scene graphs have overlapping regions to maintain global consistency. The leaf nodes $l$ are visualized as boxes with their local Cartesian coordinate axis. We also visualize the transformations and scaling between root and leaf coordinate frames using arrows with annotated transformation and scale matrices. The overall representation model is denoted as $F_{\theta}$.}
   \label{fig:progressive}
\end{figure*}
\subsection{Progressive Neural Scene Graph}

Existing dynamic NeRF systems have difficulties in large-scale urban scenes. They use a single, global representation for the entire environment, which limits their scene representation capacity. It causes problems when modeling large scenes: \textit{\textbf{a)} a single model
with fixed capacity cannot represent arbitrarily large-scale scenes. \textbf{b)} any misestimation has a global impact and might cause false reconstruction.} 

To this end, we design a progressive neural scene graph to represent city-scale scenes with multi-dynamic objects. Instead of pre-dividing the scene with the camera poses, our method dynamically adjust local scene bounds during optimization, preventing the splitting of dynamic elements when applied to dynamic environments and improving the quality of reconstructing dynamic objects.

The scene graph $\mathrm{SG}$, illustrated in Fig. \ref{fig:progressive}, is composed of cameras, 
static nodes and a set of dynamic nodes which represent the
dynamic components of the scene, including the object's appearance, shape, and classes. The entire scene is decomposed into several local scene representations:
\begin{gather}
\{I_i,P_i\}_{i=1}^M\mapsto \{\mathrm{SG}^1_{\theta_1},\mathrm{SG}^2_{\theta_2},\dots ,\mathrm{SG}^n_{\theta_n}\}\mapsto\{c,\sigma\} \notag \\ 
\mathrm{SG}=\langle\mathcal{W}, C,L,F, E\rangle
\end{gather}
$P_i$ denotes the projection of the LiDAR point cloud at frame i. The root node of scene graph denoted as $\mathcal{W}$, $ C$ is a leaf node representing the camera in the global coordinate system and its intrinsic $K \in \mathbb{R}^{3 \times 3}$. $L$ denotes the leaf node of scene graph(e.g. objects and background). $F$ represents the neural networks for both static and dynamic nodes: $ F=F_{\theta_{b c k g}} \cup\left\{F_{\theta_1}, F_{\theta_2},\dots, F_{\theta_{c}}\right\}$.
$E$ are edges that represent affine transformations
from $i$ to $i'$:
\begin{equation}
\begin{aligned}
E & =\left\{e_{i, i'} \in\left\{[], \boldsymbol{T}_{i\to i'}\right\}\right\}, \\
\text { with } \quad \boldsymbol{T}_{i\to i'} & =\left[\begin{array}{cc}
\boldsymbol{R}_{i\to i'} & \boldsymbol{t}_{i\to i'} \\
\mathbf{0} & 1
\end{array}\right],\\ \boldsymbol{R}_{i\to i'} &\in \mathrm{SO(3)},\quad \boldsymbol{t}_{i\to i'} \in \mathbb{R} {. }
\end{aligned}
\end{equation}

For all edges originating from the root node $\mathcal{W}$, we assign transformations between the global world space and the local objects with $T$. We compute the scaling factor $S_i$, assigned at edge $e_i$ between an object node and its shared representation model $F$. 

We present our progressive scheme in Algorithm~\ref{algorithm}. The algorithm describes a method for constructing a Progressive Neural Scene Graph that dynamically allocates a new local scene graph trained with frames within a temporal window, allowing us to scale up the representation to an arbitrarily large scene.

Whenever the estimated camera pose trajectory reaches the bound of the current scene graph. 
we initialize the local representation with a local subset of frames, and then we progressively introduce subsequent frames to the optimization. For each frame, it optimizes the local neural scene graph and progresses to the next frame until a predefined condition is met. Each subset contains some overlapping images, essential for achieving consistent reconstructions in the global scene.
When the limit for a local neural scene graph is reached, it initializes a new one, centered around the last pose, and continues the process until all frames are processed, contributing to the complete scene reconstruction.

For the $j$-th local scene graph, we define its spatial bound $B_j$ as an axis-aligned bounding box in the world coordinate system. Let $c_i$ denote the camera center of frame $i$, $\mathcal{F}_i(d_n,d_f)$ denote the frustum corner points between the near and far planes, and $\mathcal{O}_i$ denote the set of 3D bounding boxes of dynamic objects observed at frame $i$. The local bound is initialized and updated as
\begin{equation}
\begin{aligned}
B_j =
\mathrm{AABB}\Big(
&\{c_i\}_{i=p_j}^{q_j}
\cup \{\mathcal{F}_i(d_n,d_f)\}_{i=p_j}^{q_j} \\
&\cup \{\mathcal{O}_i\}_{i=p_j}^{q_j}
\Big)
\oplus \epsilon .
\end{aligned}
\end{equation}
where $\epsilon$ is a safety margin and $\oplus$ denotes bound expansion. During optimization, when a newly observed dynamic object intersects the boundary, we expand $B_j$ to include the complete 3D object box, avoiding the splitting of dynamic objects across local graphs. A new local scene graph is initialized when the current camera center moves outside $B_j$ or when the temporal window reaches the maximum length. Neighboring local graphs share several overlapping frames, and the overlapping scene representations are fused by inverse distance weighting.
 Whenever the camera pose leaves the bound of the current scene representation, we stop optimizing previous ones (freeze the network parameters). At this point, we can reduce memory requirements and clear any supervising frames that are not needed anymore.

The progressive scene graph effectively addresses the camera's large-scale ego-motion issue. It allows our scene reconstruction to scale to arbitrary sizes, significantly enhancing our capability to represent large-scale urban scenes. 
\noindent \textbf{Overlap and IDW fusion.}
Adjacent local scene graphs share $N_{\mathrm{ov}}$ overlapping frames to maintain cross-graph consistency. 
For the $j$-th local scene graph with frame interval $[p_j,q_j]$, the next graph starts from
\begin{equation}
p_{j+1}=q_j-N_{\mathrm{ov}}+1 .
\end{equation}
For a query point $\mathbf{x}$ located in the overlapping region of multiple local scene graphs, we fuse their predictions using inverse-distance weighting. We use 10\% of the local window length for $N_{\mathrm{ov}}$.
Let $\boldsymbol{\mu}_j$ denote the center of the spatial bound $B_j$ of the $j$-th local scene graph. 
The normalized weight of graph $j$ is computed as
\begin{equation}
w_j(\mathbf{x})=
\frac{(\|\mathbf{x}-\boldsymbol{\mu}_j\|_2+\delta)^{-1}}
{\sum_{k\in\mathcal{G}(\mathbf{x})}(\|\mathbf{x}-\boldsymbol{\mu}_k\|_2+\delta)^{-1}},
\end{equation}
where $\mathcal{G}(\mathbf{x})$ denotes the set of local scene graphs whose bounds contain $\mathbf{x}$, and $\delta$ is a small constant for numerical stability. 
The final color and density are fused with $w_j$.

\noindent \textbf{Handling long-term occlusion and object re-entry.} The proposed progressive scene graph handles short-term occlusions and adjacent local-graph transitions through overlapping frames and object-centric dynamic nodes. When the same dynamic object is observed in two neighboring local graphs, its 3D bounding boxes and track identity are used to associate the corresponding object node, while the overlapping frames provide consistency constraints. For longer-term occlusions or cases where an object leaves the field of view and re-enters after several local graphs, the current framework depends on the reliability of the external 3D detector/tracker. If the object identity is preserved, its object-centric latent representation can be reused or updated. If the identity is lost, the object is initialized as a new dynamic node. In this case, the class-shared decoder and foundation-model prior still help reconstruct the object appearance and geometry, but strict long-term identity consistency is not guaranteed. We therefore regard robust long-term re-identification across distant local graphs as an important future direction.

\begin{algorithm}[t]
\caption{Progressive Neural Scene Graph Construction}
\label{algorithm}
\begin{algorithmic}[1]
\REQUIRE Images $\{I_i\}_{i=1}^{M}$, poses $\{\mathbf{T}^{w\leftarrow c}_i\}_{i=1}^{M}$, object boxes $\{\mathcal{B}_{i,k}\}$, overlap size $O$, maximum local radius $r_{\max}$
\STATE $j \leftarrow 1$, $s_j \leftarrow 1$
\STATE Initialize local graph $\mathcal{G}_j$ with frame set $\mathcal{W}_j \leftarrow \emptyset$
\STATE Initialize local bound $\mathcal{B}_j$ using the camera center $\mathbf{o}_{s_j}$
\FOR{$i=1$ to $M$}
    \STATE Compute camera center $\mathbf{o}_i$ from $\mathbf{T}^{w\leftarrow c}_i$
    \IF{$\mathbf{o}_i \in \mathcal{B}_j$ \textbf{and} $\|\mathbf{o}_i-\mathbf{o}_{s_j}\|_2 < r_{\max}$}
        \STATE Add frame $i$ to $\mathcal{W}_j$
        \STATE Update $\mathcal{B}_j$ using $\mathbf{o}_i$ and visible object boxes $\{\mathcal{B}_{i,k}\}$
        \STATE Optimize the active graph $\mathcal{G}_j$ with frames in $\mathcal{W}_j$
    \ELSE
        \STATE Freeze $\mathcal{G}_j$
        \STATE $j \leftarrow j+1$
        \STATE $s_j \leftarrow \max(1, i-O)$
        \STATE Initialize $\mathcal{G}_j$ with overlapping frames $\{s_j,\ldots,i\}$
        \STATE Initialize $\mathcal{B}_j$ using camera centers and object boxes in $\{s_j,\ldots,i\}$
        \STATE Optimize the active graph $\mathcal{G}_j$
    \ENDIF
\ENDFOR
\STATE Fuse overlapping local graph predictions during rendering.
\end{algorithmic}
\end{algorithm}


\begin{figure}[htbp]
  \centering
  
   \includegraphics[width=\linewidth]{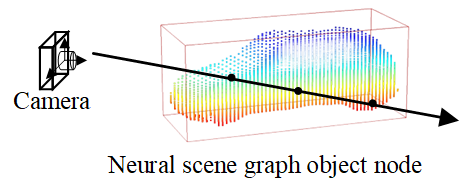}

    \caption{ Ray-Box Intersection. We use AABB ray-box sampling strategy. The boxes are now defined as an axis-aligned bounding box (AABB) with a minimum bound [-1,-1,-1] and a maximum bound [1, 1,1].}
   \label{fig:sample}

\end{figure}

\subsection{Background Model}
Our local scene graph includes a background representation node, as illustrated at the top of Fig. \ref{fig: pipeline}, which is responsible for modeling the static components of the environment. To more effectively capture scene texture, geometric details, and overall representation accuracy, we incorporate a multiresolution hash encoding method for background modeling. For a given input coordinate x, the surrounding voxels are identified across L resolution levels, and indices are assigned to their corner vertices by hashing the corresponding integer coordinates. The associated F dimensional feature vectors are then retrieved from the hash tables and linearly interpolated according to the relative position of x within each voxel. The interpolated features from all resolution levels are concatenated to form the final scene feature representation. The voxel resolutions are selected as a geometric progression ranging from the coarsest to the finest level, enabling efficient representation of both global structure and fine scale details $N_{min},N_{max}$:
\begin{equation}
\begin{aligned}
N_l & :=\left\lfloor N_{\min } \cdot b^l\right\rfloor, \\
b & :=\exp \left(\frac{\ln N_{\max }-\ln N_{\min }}{L-1}\right)
\end{aligned}
\end{equation}
$N_{max}$is chosen to match the finest detail in the training data.  We use spatial hash function:
\begin{equation}
h(\mathbf{x})=\left(\bigoplus_{i=1}^d x_i \pi_i\right) \quad \bmod T
\end{equation}
where $\bigoplus$  denotes the bit-wise XOR operation and $\pi_3$ are large prime numbers. Effectively, this formulation applies an XOR operation to the outputs of per dimension linear congruential pseudo random permutations, thereby decorrelating the influence of individual dimensions on the resulting hash value. We choose $\pi_1:=1$ for better cache coherence, $\pi_2=2654435761$, and $\pi_3=805459861$.

Unlike the original NeRF, which applies Fourier positional encoding to 3D positions, our background branch uses multiresolution hash encoding for spatial coordinates. We only apply Fourier encoding to the viewing direction:
\begin{equation}
\begin{aligned}
\gamma_d(\mathbf{d}) =
\Big[
&\mathbf{d},
\sin(2^0\pi\mathbf{d}), \cos(2^0\pi\mathbf{d}),
\ldots, \\
&\sin(2^{K_d-1}\pi\mathbf{d}), 
\cos(2^{K_d-1}\pi\mathbf{d})
\Big].
\end{aligned}
\end{equation}

The background decoder follows a two-stage design. The first MLP predicts volume density and a geometry feature from the hash-encoded position:
\begin{equation}
\sigma_b(\mathbf{x}),\mathbf{z}_b(\mathbf{x})
=
F_{\theta_{bkg1}}
\left(
\mathbf{f}_{\mathrm{hash}}(\mathbf{x})
\right).
\end{equation}
The second MLP predicts view-dependent color using the geometry feature and the encoded viewing direction:
\begin{equation}
\mathbf{c}_b(\mathbf{x},\mathbf{d})
=
F_{\theta_{bkg2}}
\left(
\mathbf{z}_b(\mathbf{x}), \gamma_d(\mathbf{d})
\right).
\end{equation}

\subsection{Foundation-Guided Dynamic Object Representation}

All existing NeRF methods have difficulties in multi-dynamic objects in urban scenes. The original NeRF method assumes the
density $\sigma$ is static, but both the density $\sigma(\boldsymbol{x})$ and color $\boldsymbol{c}(\boldsymbol{x})$ depend on time (and video). SUDS~\cite{suds} and Emernerf~\cite{emernerf} use an optical flow method to differentiate dynamic objects from the background. However, their method cannot model dynamic objects effectively, resulting in blurred object portions and numerous artifacts in the rendered images. Furthermore, most of the dynamic objects only appear in a few observations. This setting starkly contrasts the majority of existing works
that observe multiple views of the same object. To address this challenging problem, we propose a frequency auto-encoder in dynamic objects node to learn a normalized, object-centric high-level feature that describes and disentangles shape, appearance, and pose.

Auto-encoder is a type of neural network architecture used for unsupervised learning of efficient encodings of input data. They consist of an encoder network that maps the input to a latent code, and a decoder network that reconstructs the input from the code. We employ a vision foundation model as the encoder to extract accurate shape and appearance representations, benefiting from the strong priors learned through large scale pretraining. This design enables the proposed method to more effectively address the \textbf{sparse observation} challenges of dynamic objects in autonomous driving scenarios. The code is a compressed representation that encodes the most salient features of the data.

 \noindent\textbf{Dynamic Objects Node $L_o$.} The architecture of our dynamic branch is shown in Fig. \ref{fig: pipeline}. We consider each rigid part that changes its position through the observation as the dynamics. Each dynamic object is represented by the frequency auto-encoder in the local space of its node $L_o$ and position $p_o$.  As the global location $p_o$ of a
dynamic object moves between frames, its scene representation also moves. We introduce local 
 Object Coordinate Space (OCS) $F_o$, fixed and aligned with the object’s pose. The transformation in the the global frame $F_W$ to $F_o$ is given by
\begin{equation}
\boldsymbol{x}_o=\boldsymbol{S}_o \boldsymbol{T}_o^w \boldsymbol{x} 
\end{equation}
where $S_o$ is a matrix with the inverse length values in the middle. 
$s_o = [L_o, H_o, W_o]$ with $S_o = diag(1/s_o)$. The scaling factor allows the network to learn size-independent representation. We share the weights of the network for the objects of the same class (such as cars, vans and trucks that were categorized during preprocessing.)

\noindent \textbf{Shape and Appearance Encoder.} For a input image $I$, we use the segment anything model and 3D object detector to get the 2D mask $M$ and 3D bounding box $\beta$. We encode the input images $I$ depicting a given object of interest into a shape code $\mathcal{S}_\theta$ and an appearance code $\mathcal{A}_\theta$ via a neural network. Inspired by \cite{autorf}, we design a two-head output architecture with the vision foundation extractor for generating the shape code $l_s$ and appearance code $l_a$, respectively.

We use the DINOv2 backbone to extract image features. The output tokens of the DINO model consist of both a class token and patch tokens. The class token primarily encodes high-level semantic information, which we map through a lightweight MLP to generate the shape code of the dynamic object. The patch tokens mainly capture local appearance cues such as color and texture. We apply a mean pooling operation over the patch tokens, followed by another lightweight MLP, to obtain the appearance code. The final latent representation comprises both the shape code and appearance code, each with 128 dimensions. Image inputs are scaled to a maximum of 320 pixels in each dimension, maintaining the original aspect ratio.


\noindent \textbf{Shape and Appearance Decoder $\phi$.} The shape code $l_s$ is fed to the shape decoder
network $\phi_s$ which implicitly outputs an density value $\sigma$ of the given coordinate$x$. Unlike the shape code decoder, the appearance code decoder $\phi_a$ uses both shape and appearance codes as the input with the positional and directional information. The appearance decoder also takes the dynamic object node information to perceive the object's position better and implicitly outputs the RGB color $c$.  

\noindent\textbf{Shape Decoder.} The shape decoder is a MLP that is made of 5 ResNet blocks with hidden dimension 128. At each layer, we feed the previous feature map and the positional encoding of the query points. In order to match the dimensionality of the positional encoding with the hidden dimensions of the MLP, we apply a single linear layer and aggregate the output with the intermediate feature maps by a simple per-channel mean pooling.

\noindent\textbf{Color Decoder.} For decoding the color, we use a similar architecture as for the shape decoder: A MLP of 5 ResNet blocks with hidden dimensionality of 128 and additional linear aggregations for additional input: As for the shape decoder, we aggregate intermediate features with positional encodings by mean pooling. Furthermore, on the third layer we pass in the same way the output of the corresponding layer of the shape encoder. This enables the color decoder to incorporate estimated shape information. On the final two layers, we pass the view direction (encoded as 3-dimensional vector) to account for view-dependent effects.

\noindent \textbf{Frequency Modulation.} 
In real-world urban scenes, dynamic objects only appear in a few observations, which significantly differs from the settings of the original NeRF method. In sparse view settings, the Neural radiance field is prone to overfitting to these 2D images without explaining 3D geometry in a multi-view consistent way, leading to reconstruction failure. Prior work~\cite{freenerf} has already demonstrated that this issue
is presumably caused by the high-frequency feature of the input. 

To this end, we further improve the feature-extracting foundation model with a frequency regularization method. The original method addressed the sparsity issue by utilizing a CNN network to encode the high-frequency features, which leads to the overfitting of high-frequency features. 
In this work, we use a linearly increasing frequency mask $\boldsymbol{\alpha}$ to regulate the visible frequency spectrum of the position embedding function:
\begin{gather}
\gamma_L(\mathbf{x})=\left[\sin (\mathbf{x}), \cos (\mathbf{x}), \ldots, \sin \left(2^{L-1} \mathbf{x}\right), \cos \left(2^{L-1} \mathbf{x}\right)\right] \notag\\
\gamma_L^{\prime}(t, T ; \mathbf{x})=\gamma_L(\mathbf{x}) \odot \boldsymbol{\alpha}(t, T, L)
\end{gather}

where $t$ and $T$ are the current training iteration and the final iteration of frequency regularization. The mask $\boldsymbol{\alpha}$ is increased based on the training time:
\begin{equation}
\boldsymbol{\alpha}_n(t, T, L)= \begin{cases}1 & \text { if } n \leq \frac{t \cdot L}{T}+3 \\ \frac{t \cdot L}{T}-\left\lfloor\frac{t \cdot L}{T}\right\rfloor & \text { if } \frac{t \cdot L}{T}+3<n \leq \frac{t \cdot L}{T}+6 \\ 0 & \text { if } n>\frac{t \cdot L}{T}+6\end{cases}
\end{equation}
$\boldsymbol{\alpha}_n$ denotes the n-th value of $\boldsymbol{\alpha}$. Our method starts with raw inputs without positional encoding (low-frequency). Then, our mask linearly increases the visible frequency brand as training progresses.

The constants $+3$ and $+6$ define a smooth transition band for progressive frequency activation. 
The offset $+3$ keeps a small number of low-frequency components active from the early training stage, allowing the network to first capture coarse object geometry and appearance. 
The interval between $+3$ and $+6$ serves as a soft annealing region, where the next frequency bands are gradually introduced instead of being activated abruptly. 
This design prevents unstable high-frequency signals from dominating the early optimization stage, while still allowing high-frequency details to be learned in later iterations. We gradually provide our network with high-frequency signals to avoid over-smoothness.

Frequency regularization restricts the high-frequency components of the neural network weights or activations during training. This prevents the model from overfitting to noisy high-frequency patterns in the data.
It acts as a low-pass filter, smoothing the weight updates and forcing the model to learn more general, low-frequency representations, ultimately improving generalization. All the additional information related to scene graph dynamic objects nodes: the coordinate $\boldsymbol{x}$, direction $\theta \phi$, and object pose is embedded by our frequency encoding before they are put into the auto-encoder.

Our method addresses the reconstruction of dynamic objects from sparse observations, distinct from prior approaches such as FreeNeRF \cite{freenerf} that exclusively focus on static objects. Unlike FreeNeRF \cite{freenerf}, which encodes only input points, our frequency auto-encoder architecture incorporates dynamic object poses, input points, and appearance and shape features from scene graph nodes. This adaptation significantly enhances the model's capacity to handle complex dynamic scenes, surpassing the capabilities of the previous method. Our frequency auto-encoder significantly improves the scene representation of dynamic objects and achieves great performance in image reconstruction, novel view synthesis, and object editing.
\begin{figure*}[t]
  \centering
   
   \includegraphics[width=\linewidth]{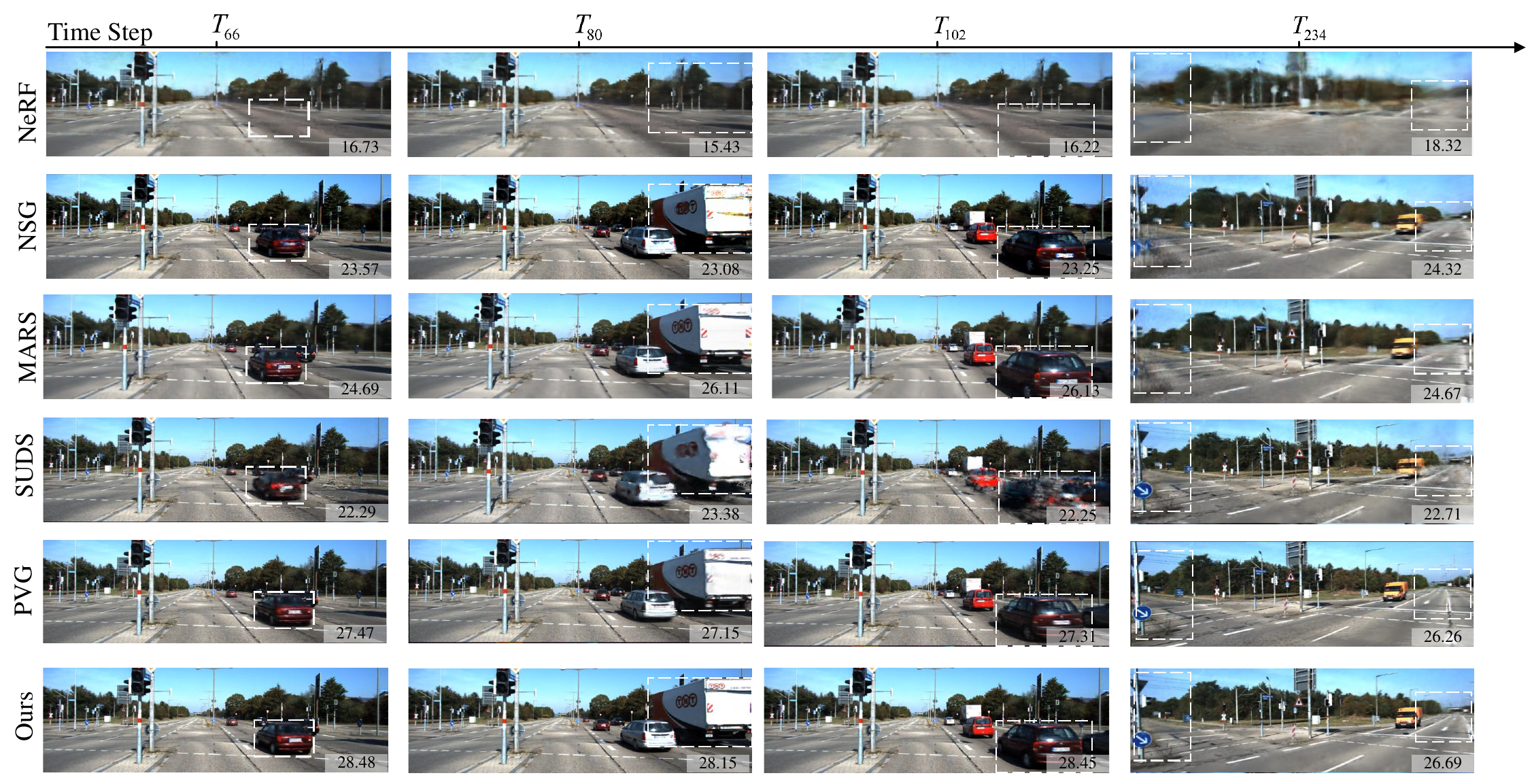}
    \caption{Qualitative results on reconstruction and novel scene arrangements of a scene from the KITTI dataset \cite{kitti} for NeRF \cite{NeRF}, NSG \cite{nsg}, SUDS \cite{suds}, MARS~\cite{mars}, PVG~\cite{pvg} and our method. From left to right, these images correspond to different timesteps captured in the dynamic scene. We place the PSNR values of each scene in the bottom right corner. NeRF and SUDS are limited in their ability to properly represent the dynamic parts of the scene. In contrast, our neural scene graph method achieves high-quality view synthesis results for both reconstruction and novel scene synthesis, regardless of scene dynamics.}
   \label{fig: kitti}
\end{figure*}

\begin{table*}
\centering
\caption{Image reconstruction and novel view synthesis results on the complete KITTI~\cite{kitti} dataset. We compare our method with others on PSNR, SSIM, and LPIPS metrics. For PSNR and SSIM, higher is better; for LPIPS, lower
is better. Our method generates higher-quality rendering images both in image reconstruction and novel view synthesis.}
\scalebox{0.75}{
\setlength{\tabcolsep}{0.5mm}{
\begin{tabular}{lccccccccccccccc}
\toprule
\multirow{3}{*}{Methods} & \multirow{3}{*}{\begin{tabular}[c]{@{}c@{}}Object\\ Manipulation\end{tabular}} & \multirow{3}{*}{\begin{tabular}[c]{@{}c@{}}Ego-\\ Motion\end{tabular}} & \multicolumn{3}{c}{Image Reconstruction}                                                            & \multicolumn{9}{c}{Novel View Synthesis}                                                                                                                                                                                                                                                                        \\ 
                         &                                                                                &                                                                        & \multicolumn{3}{c}{KITTI \cite{kitti}}                                                                           & \multicolumn{3}{c}{KITTI - 75\%}                                                                    & \multicolumn{3}{c}{KITTI - 50\%}                                                                    & \multicolumn{3}{c}{KITTI - 25\%}                                                                    \\ 
                         &                         &                        & PSNR$\uparrow$                  & SSIM$\uparrow$                  & LPIPS$\downarrow$               & PSNR$\uparrow$                  & SSIM$\uparrow$                  & LPIPS$\downarrow$               & PSNR$\uparrow$                  & SSIM$\uparrow$                  & LPIPS$\downarrow$               & PSNR$\uparrow$                  & SSIM$\uparrow$                  & LPIPS$\downarrow$               \\ \hline
NeRF \cite{NeRF}                    &   \color{BrickRed}{\xmark}                      &  \color{BrickRed}{\xmark}                      & 16.98          & 0.509          & 0.524          & 16.81          & 0.509          & 0.524          & 16.64          & 0.504          & 0.535          & 16.55          & 0.498          & 0.544                           \\
NeRF$+$Time              & \color{BrickRed}{\xmark}                        & \color{BrickRed}{\xmark}                       & 24.18          & 0.677          & 0.237          & 22.05              & 0.612              & 0.262              & 21.90              & 0.602              & 0.239              & 20.17              & 0.591              & 0.253                           \\
NSG \cite{nsg}                     & \color{ForestGreen}{\cmark}                        & \color{BrickRed}{\xmark}                        & 26.66          & 0.857          & 0.071          & 25.25          & 0.801          & 0.089          & 24.81          & 0.786          & 0.105          & 22.31          & 0.752          & 0.134                            \\
SUDS \cite{suds}                    &   \color{BrickRed}{\xmark}                       &  \color{ForestGreen}{\cmark}                        & 28.31          & 0.857          & 0.065          & 25.21          & 0.834          & 0.144          & 24.29          & 0.821          & 0.156          & 19.73          & 0.758          & 0.217                            \\
S-NeRF~\cite{snerf}              & \color{BrickRed}{\xmark}                        & \color{ForestGreen}{\cmark}                       & 19.25 & 0.664 & 0.193 & 18.71 & 0.606 & 0.352 & 18.31 & 0.598 & 0.361 & 17.93 & 0.586 & 0.365                  \\
3DGS~\cite{3dgs}  & \color{BrickRed}{\xmark}                        & \color{ForestGreen}{\cmark} &  21.02 & 0.811 & 0.202 & 19.54 & 0.776 & 0.224 & 18.95 & 0.763 & 0.232 & 18.27 & 0.758 & 0.236 \\
Mars~\cite{mars}  & \color{ForestGreen}{\cmark}                        & \color{BrickRed}{\xmark}           & 27.96 & 0.900 & 0.105 & 24.31 & 0.845 & 0.102 & 24.19 & 0.821 & 0.118 & 22.85 & 0.732 & 0.125 \\
EmerNeRF \cite{emernerf}                    &   \color{BrickRed}{\xmark}                       &  \color{ForestGreen}{\cmark}                       & 29.40 & 0.921 & 0.061 & 26.32 & 0.862 & 0.089 & 25.02 & 0.851 & 0.092 & 22.98 & 0.819 & 0.122                         \\
StreetGaussian~\cite{streetgaussian} & \color{BrickRed}{\xmark}                        & \color{ForestGreen}{\cmark}  & 29.45 & 0.926 & 0.059 & 26.79 & 0.844 & 0.083 & 25.31 & 0.841 & 0.089 & 24.54 & 0.828 & 0.098 \\
PVG~\cite{pvg} & \color{BrickRed}{\xmark}                        & \color{ForestGreen}{\cmark}  & 29.37 & 0.922 & 0.061 & 26.43 & 0.839 & 0.086 & 25.12 & 0.837 & 0.091 & 24.23 & 0.823 & 0.101 \\
Ours                     &  \color{ForestGreen}{\cmark}                       & \color{ForestGreen}{\cmark}                       & \textbf{30.31} & \textbf{0.931} & \textbf{0.057} & \textbf{27.75} & \textbf{0.887} & \textbf{0.080} & \textbf{26.55} & \textbf{0.878} & \textbf{0.086} & \textbf{25.69} & \textbf{0.856} & \textbf{0.091} \\ \bottomrule
\end{tabular}}}

\label{tab:kitti}

\end{table*}

\begin{table*}
\centering
\caption{We report the PSNR, SSIM, and LPIPS metrics results on  VKITTI2~\cite{vkitti} dataset. Our approach outperforms prior methods of reconstruction and novel view synthesis across all evaluation metrics.}
\scalebox{0.82}{
\setlength{\tabcolsep}{0.5mm}{
\begin{tabular}{lccccccccccccc}
\toprule
\multirow{3}{*}{Methods} &\multirow{3}{*}{Year} & \multicolumn{3}{c}{Image Reconstruction}         & \multicolumn{9}{c}{Novel View Synthesis}                                                                                                               \\  
                         & & \multicolumn{3}{c}{VKITTI2 \cite{vkitti}}                       & \multicolumn{3}{c}{VKITTI2 - 75\%}                & \multicolumn{3}{c}{VKITTI2 - 50\%}                & \multicolumn{3}{c}{VKITTI2 - 25\%}                \\ 
                         & & PSNR$\uparrow$           & SSIM$\uparrow$           & LPIPS$\downarrow$          & PSNR$\uparrow$           & SSIM$\uparrow$           & LPIPS$\downarrow$          & PSNR$\uparrow$           & SSIM$\uparrow$           & LPIPS          & PSNR$\uparrow$           & SSIM$\uparrow$           & LPIPS$\downarrow$          \\ \hline
NeRF \cite{NeRF}      & ECCV'20               & 17.63          & 0.525          & 0.511          & 17.11          & 0.513          & 0.514          & 16.94          & 0.517          & 0.521          & 16.65          & 0.504          & 0.537          \\
NeRF + Time    &          & 24.54          & 0.683          & 0.235          & 22.59              & 0.619              & 0.252              & 22.15              & 0.612              & 0.261              & 21.82              & 0.615              & 0.233              \\
NSG~\cite{nsg}  & CVPR'21                   & 27.18         & 0.894          & 0.092          & 25.45          & 0.886          & 0.094          & 24.95          & 0.794          & 0.101          & 23.49          & 0.758          & 0.105          \\
SUDS \cite{suds}    & CVPR'23                 & 28.34          & 0.887          & 0.049          & 27.52          & 0.864          & 0.064          & 26.25          & 0.845          & 0.074          & 21.47          & 0.788          & 0.088          \\
S-NeRF~\cite{snerf} & ICLR'23  & 20.53 & 0.874 & 0.181 & 19.46 & 0.646 & 0.342 & 19.11 & 0.604 & 0.354 & 18.43 & 0.591 & 0.359 \\

StreetSurf~\cite{streetsurf} & Arxiv'23 & 24.89 & 0.839  & 0.242 & 23.88  & 0.785  & 0.273 & 22.79 & 0.776 & 0.289 & 21.59 & 0.746 & 0.303 \\

3DGS~\cite{3dgs} & TOG'24 & 22.34 & 0.868  & 0.179 & 21.42  & 0.657  & 0.339 & 21.11 & 0.624 & 0.347 & 20.84 & 0.863 & 0.329 \\

Mars~\cite{mars} & CICAI'23  & 30.26 & 0.925 & 0.083 & 29.79 & 0.917 & 0.088 & 29.63 & 0.916 & 0.087 & 27.01 & 0.877 & 0.104 \\
Streetgaussian~\cite{streetgaussian}  & ECCV'24  & 31.37 & 0.928 & 0.081 & 30.08 & 0.922 & 0.084 & 30.03 & 0.910 & 0.089 & 28.13 & 0.905 & 0.099 \\
PVG~\cite{pvg}  & IJCV'26  & 31.69 & 0.931 & 0.077 & 30.45 & 0.925 & 0.080 & 30.15 & 0.914 & 0.086 & 28.45 & 0.910 & 0.095 \\

Ours    &                 & \textbf{32.42} & \textbf{0.938} & \textbf{0.075} & \textbf{31.58} & \textbf{0.935} & \textbf{0.079} & \textbf{30.95} & \textbf{0.931} & \textbf{0.084} & \textbf{29.69} & \textbf{0.927} & \textbf{0.088} \\ \bottomrule
\end{tabular}}}

\label{tab:vkitti}
\end{table*}

\begin{figure*}[t]
  \centering
   \includegraphics[width=0.9\linewidth]{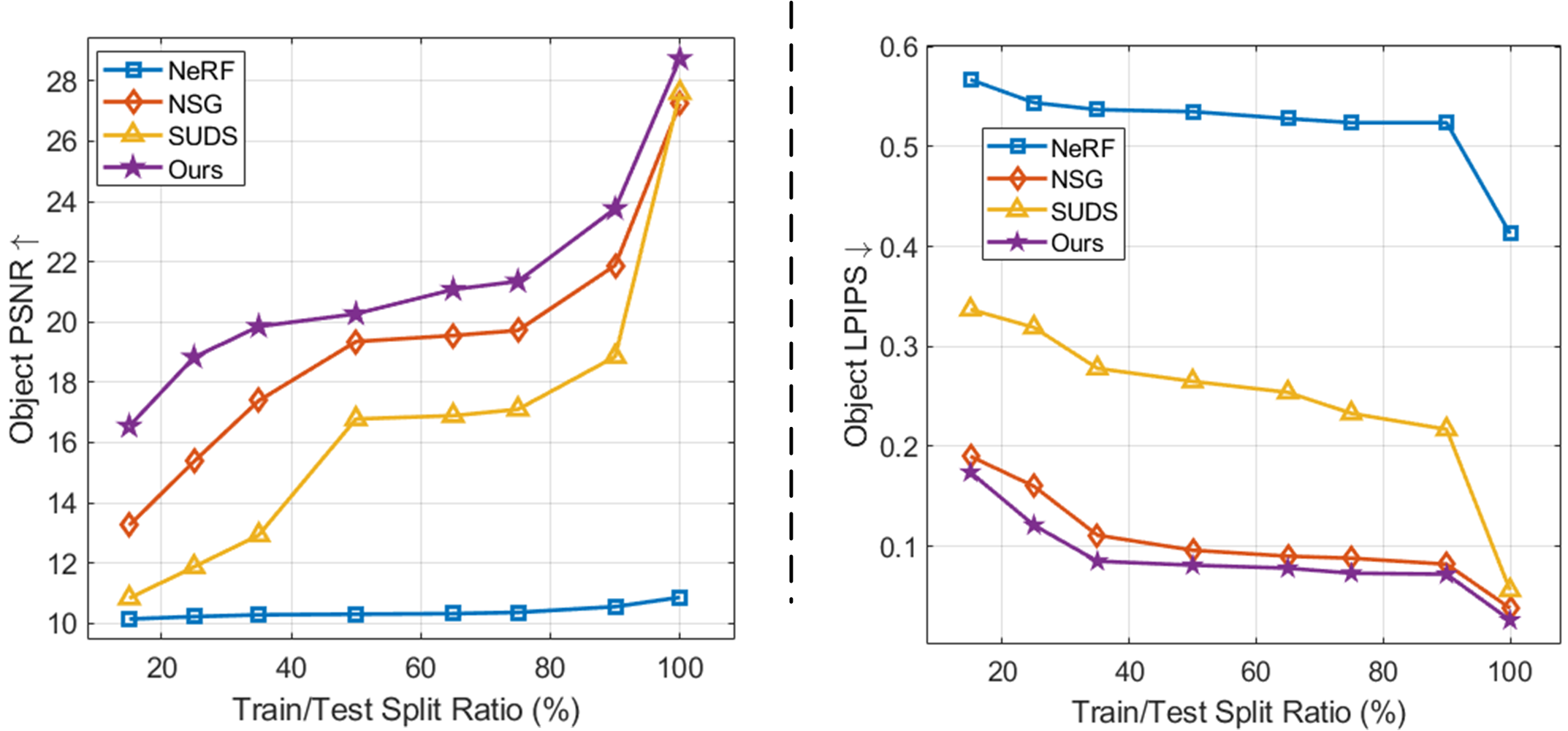}

    \caption{Dynamic objects view synthesis evaluation: measures of image fidelity plotted against train/test split ratio.}
   \label{fig:manipulation3}

\end{figure*}

\subsection{Scene Graph Rendering}
This section presents the differentiable rendering of our proposed progressive neural scene graph. Each local scene graph has a subset of overlapping images $I_k \in \mathbb{R}^{H \times W \times 3}$. The rays are generated from the images in scene graph $SG$ with the camera intrinsic $K$ and the transformation matrix $T$. We can determine a ray $r(t)=o+td$, whose origin is at the camera center of projection $o$, and ray direction $r$. 

We use different sampling strategies for the background and dynamic objects. For the background, we use a multi-plane sampling strategy to increase efficiency. We define $N_s$ planes parallel to the image plane of the initial camera pose $T_{wc}$. The sample planes are within the near $d_n$ and far planes $d_f$ with stratified distribution. For
a ray $r$, we calculate the intersections $t_{n=1}^{N_s}$ with each plane and get the sample point $\boldsymbol{x}_{bkg}$ of the background.

\noindent\textbf{Ray-Plane Intersection} \quad  Ray-Plane Intersection focuses on efficiently modeling static components using sparse planes. These planes are uniformly distributed along the scene's depth, intersecting rays to generate sampling points for the static background. In our experiment we use 10 to 32 planes to capture different movement scenarios and camera perspectives.

\noindent\textbf{Ray-Box Sampling.} For each ray $r$ in the dynamic object node, we use the ray-box sampling method. We check the rays that intersect with the dynamic object nodes and translate them to the object coordinate space. Then, we apply the efficient AABB intersection method \cite{aabb} to compute all ray-box entrance and exit points $(t_{o1}, t_{oN})$. We stratified sample points $\boldsymbol{x}_{o}$ between the entrance and exit points:
\begin{equation}
t_{on}=\frac{n-1}{N_d-1}\left(t_{oN}-t_{o1}\right)+t_{o1}
\end{equation}

Ray-Box Intersection involves transforming rays into local frames of dynamic objects, enabling computation of intersections within scaled axis-aligned bounding boxes (AABBs), as shown in Fig. \ref{fig:sample}. Equidistant quadrature points along the rays within the bounding box are sampled to represent object surfaces accurately. Experimentally, using 7 equidistant points per ray and object node was found to balance accuracy and rendering speed.

\noindent\textbf{Volume Rendering.}
The sample points along the rays can be represented as:$\left\{t_i\right\}_{i=1}^{N_s+m_j N_d}$. $m_o$ represents the number of dynamic nodes that the ray intersects along its path. 
The color $c(x_i)$ and volumetric density $\sigma(x_i)$ of each intersection point is predicted from the respective network in the static background node $F_{\theta_{bkg}}$ or dynamic objects node $F_{\theta_{o}}$.
Then, we define the termination probability and color as:
\begin{gather}
\hat{\boldsymbol{C}}(\boldsymbol{x})=\sum_{i=1}^{N_s+m_j N_d} T_i \alpha_i \boldsymbol{c}_i, \hat{\boldsymbol{D}}(\boldsymbol{x})=\sum_{i=1}^{N_s+m_j N_d} T_i \alpha_i \delta_i
 \\
T_i=\exp \left(-\sum_{k=1}^{i-1} \sigma_k \delta_k\right), \alpha_i=1-\exp \left(-\sigma_i \delta_i\right)
\end{gather}
where $\delta_i = t_{i+1}- t_{i}$ is the distance between adjacent points.

 \begin{figure*}[h]
  \centering
    
    \includegraphics[width=\linewidth]{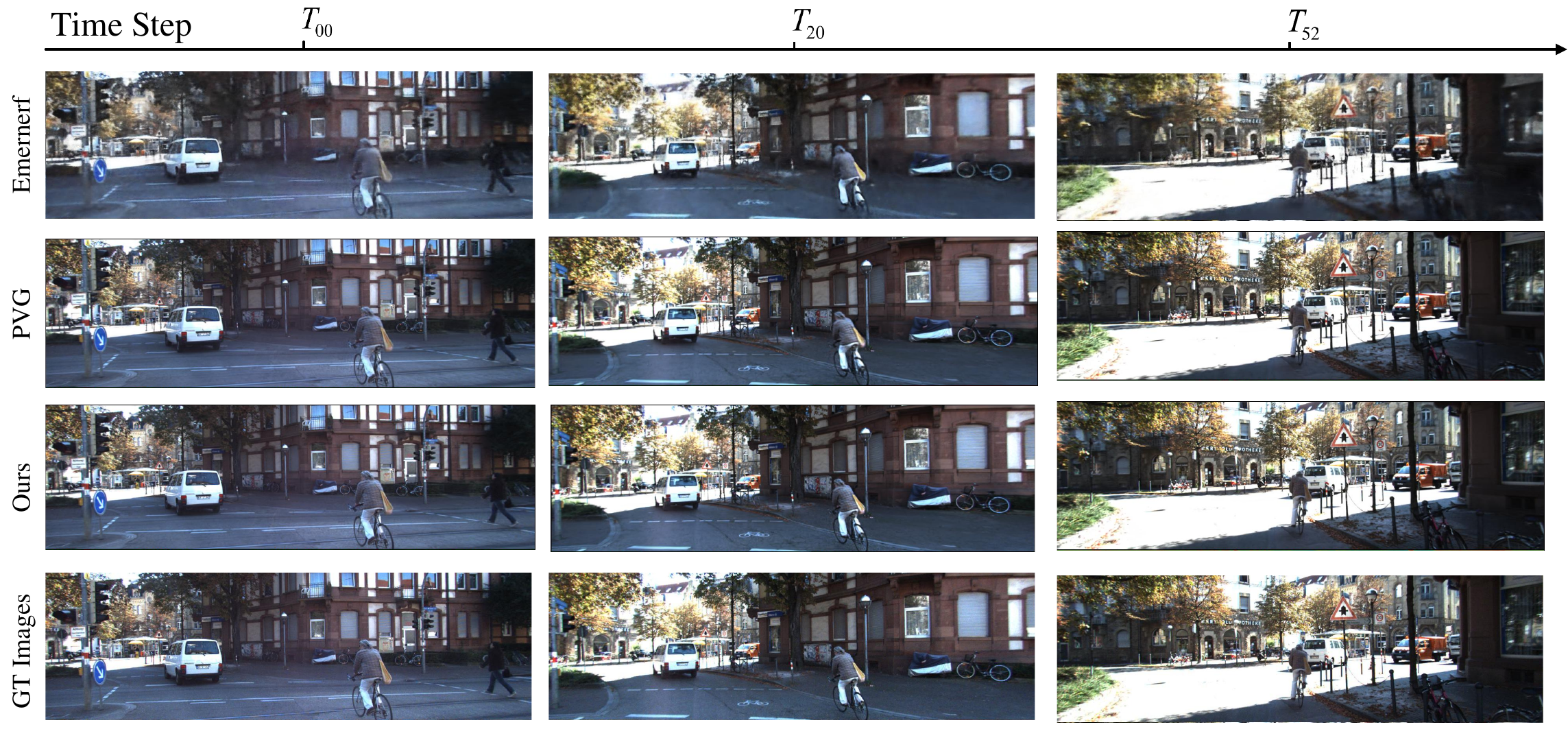}
    \caption{ We also present our method for novel view synthesis performance on the KITTI 0000 sequence, with high-dynamic human. The results clearly show that our approach achieve high accuracy reconstruction perfromance.}
   \label{fig:kitti0}
\end{figure*}

\begin{figure*}[t]
  \centering
  
   \includegraphics[width=\linewidth]{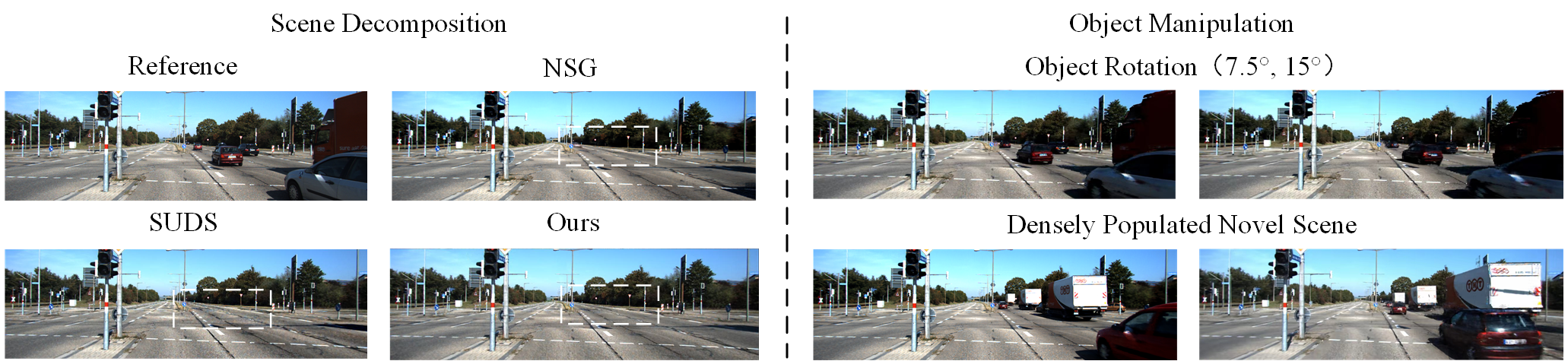}
    \caption{Scene decomposition and object manipulation. We demonstrate the decomposition capability of our progressive scene graph compared to other methods, emphasizing its capacity to isolate foreground objects and generate cleaner backgrounds. Additionally, the right part showcases our proficiency in scene editing, depicting our capability to render rotated objects and synthesize densely populated novel scenes realistically. }
   \label{fig: manipulation}
\end{figure*}

 \begin{figure*}[h]
  \centering
    
    \includegraphics[width=\linewidth]{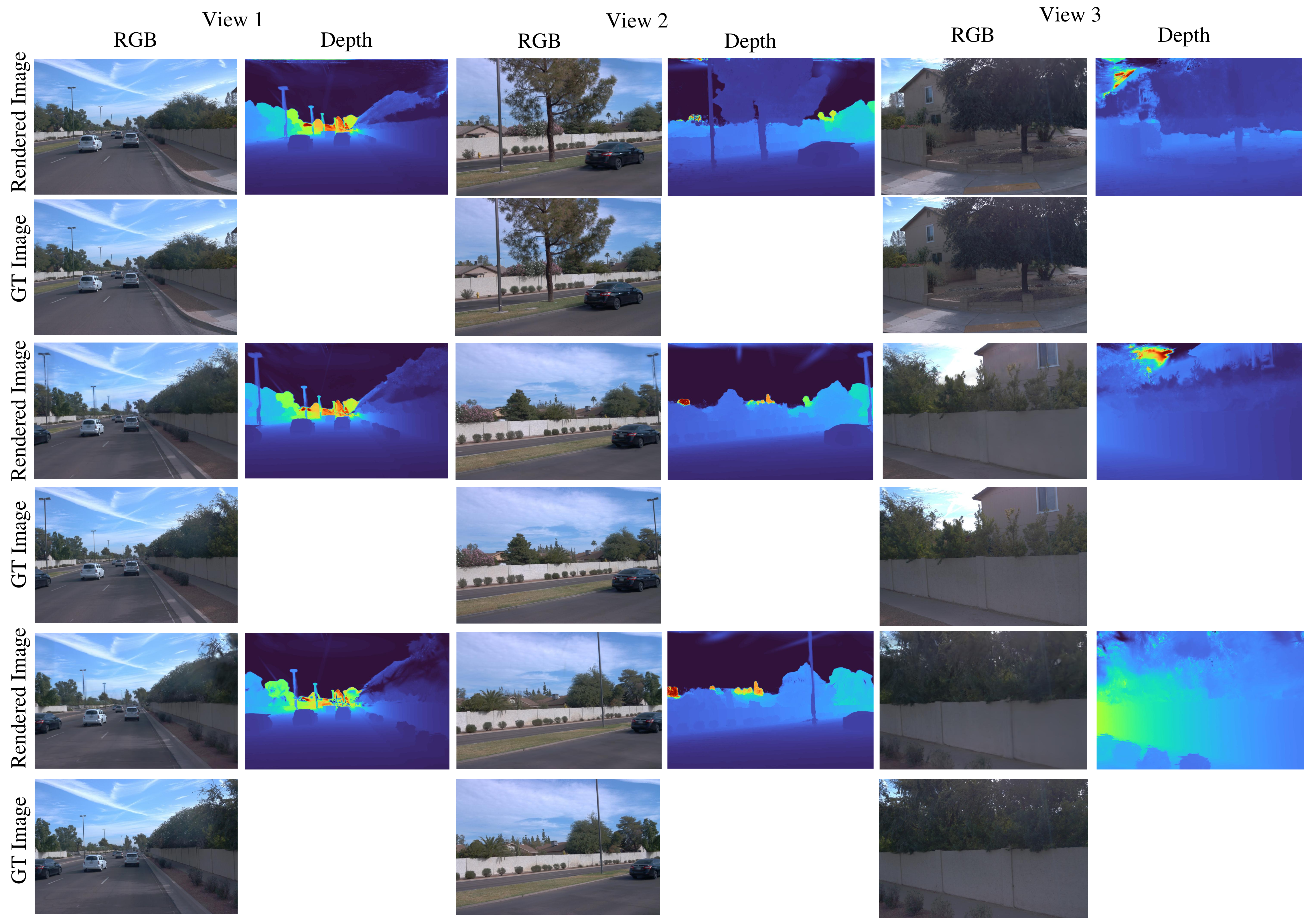}

    \caption{ We also present our method for scene reconstruction performance. The results clearly show that our approach achieve high accuracy reconstruction perfromance.}
   \label{fig:nuScenes}

\end{figure*}

 \begin{figure*}[h]
  \centering
    
    \includegraphics[width=\linewidth]{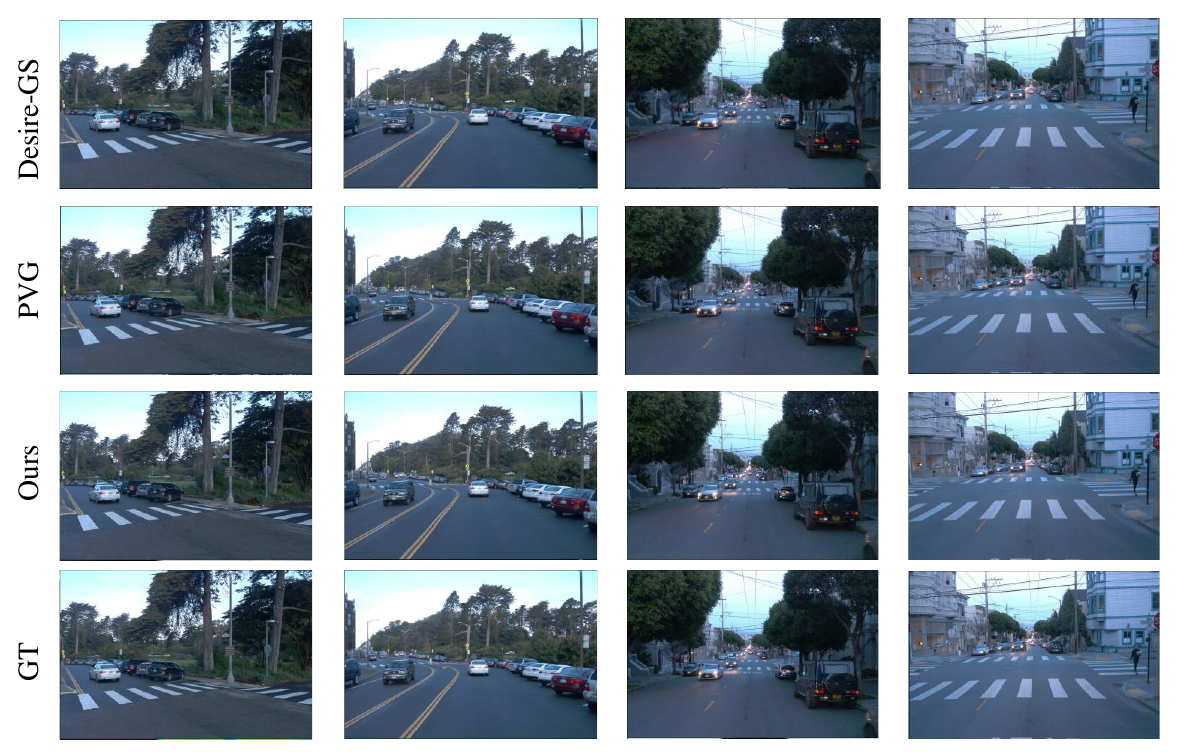}

    \caption{ We also present our method for scene reconstruction performance on the Waymo dataset~\cite{waymo}, compared with PVG and Desire-gs. The results clearly show that our approach achieve high accuracy reconstruction perfromance.}
   \label{fig:waymo}

\end{figure*}

\begin{table*}[]

\caption{Reconstruction performance on Waymo~\cite{waymo} dataset. We outperform other works on image reconstruction and novel view synthesis.}
\centering
\begin{tabular}{lllllll}
\toprule
\multirow{3}{*}{Methods} & \multicolumn{6}{c}{Waymo~\cite{waymo}}                                                           \\  
                         & \multicolumn{3}{c}{Image Reconstruction} & \multicolumn{3}{c}{Novel View Synthesis} \\
                         & PSNR$\uparrow$         & SSIM$\uparrow$        & LPIPS$\downarrow$       & PSNR$\uparrow$         & SSIM$\uparrow$        & LPIPS$\downarrow$        \\ \midrule
NeRF~\cite{NeRF}                     & 18.89        & 0.515       & 0.398       & 18.45        & 0.508       & 0.405       \\
NSG~\cite{nsg}                      & 24.08        & 0.656       & 0.441       & 21.05        & 0.575       & 0.487       \\
SUDS~\cite{suds}                    & 28.83        & 0.805       & 0.317       & 25.36        & 0.783       & 0.384       \\
S-NeRF~\cite{snerf}                   & 19.67        & 0.528       & 0.387       & 19.25        & 0.517       & 0.402       \\
StreetSurf~\cite{streetsurf}               & 26.73        & 0.849       & 0.372       & 23.79        & 0.825       & 0.401       \\
3DGS~\cite{3dgs}                     & 27.99        & 0.866       & 0.293       & 25.08        & 0.822       & 0.319       \\
MARS~\cite{mars}                     & 21.84        & 0.682       & 0.431       & 20.69        & 0.637       & 0.453       \\
EmerNeRF~\cite{emernerf}                     & 28.11        & 0.784       & 0.375      & 25.94        & 0.765       & 0.383       \\
PVG~\cite{pvg}   & 31.45        & 0.897       & 0.243      & 27.49        & 0.838       & 0.303       \\
DeSiRe-GS~\cite{desiregs}   & 31.89        & 0.904       & 0.238      & 28.84        & 0.842       & 0.300       \\
Ours                     &  \textbf{32.55}            & \textbf{0.912}            &  \textbf{0.235}           &  \textbf{29.89}            &  \textbf{0.861}           &  \textbf{0.275} \\ \bottomrule          
\end{tabular}
\label{tab:waymo}
\end{table*}

\begin{table}[!t]
\centering
\caption{Quantitative comparison with existing methods on the nuScenes dataset~\cite{nuScenes}. We report PSNR, SSIM, and LPIPS. Higher PSNR and SSIM indicate better performance, while lower LPIPS is better.}
\label{tab:comparison}
\begin{tabular}{lcccc}
\toprule
\textbf{Methods}  & \textbf{PSNR$\uparrow$} & \textbf{SSIM$\uparrow$} & \textbf{LPIPS$\downarrow$} \\
\midrule
Instant-NGP~\cite{instantngp}  & 16.78 & 0.519 & 0.570 \\
NeRF+Time  & 17.54 & 0.565 & 0.532 \\
NSG~\cite{nsg}  & 21.67 & 0.671 & 0.424 \\
Mip-NeRF~\cite{mipnerf}  & 18.08 & 0.572 & 0.551 \\
Mip-NeRF360~\cite{mipnerf360}  & 22.61 & 0.688 & 0.395 \\
Urban-NeRF~\cite{urbannerf}  & 20.75 & 0.627 & 0.480 \\
S-NeRF~\cite{snerf}  & 25.43 & 0.730 & 0.302 \\
SUDS~\cite{suds}  & 21.26 & 0.603 & 0.466 \\
EmerNeRF~\cite{emernerf}  & 26.75 & 0.760 & 0.311 \\
3DGS~\cite{3dgs}  & 26.08 & 0.717 & 0.298 \\
4DGS~\cite{4dgs}  & 19.79 & 0.622 & 0.473 \\
DrivingGaussian~\cite{drivinggaussian}  & 28.74 & 0.865 & 0.237 \\
Ours  & \textbf{29.24} & \textbf{0.878} & \textbf{0.223} \\
\bottomrule
\end{tabular}
\label{tab:nuscenes}
\end{table}

\subsection{Progressive Scene Graph Joint Learning}

For the large-scale dynamic scene, we divide the entire scene into multi-local scene graphs. Whenever the camera moves to the bound of the current scene graph, we dynamically initialize a new neural scene graph and stop optimizing the previous ones. 

When encountering the same objects observed in two local scenes, we update only the dynamic object network in the current scene graph because the optimization of the previous scene graphs is frozen. To mitigate potential quality reduction in the former scene graph caused by this strategy, each subset contains some overlapping images, essential for achieving consistent reconstructions in the global scene. We also dynamically adjust local scene bounds during optimization. This helps prevent the undesired splitting of dynamic elements and optimizes the overall reconstruction process.

In each local scene graph, we use color loss to optimize the representation model for the background and dynamic object node:
\begin{equation}
\mathcal{L}_{c}=\frac{1}{N} \sum_{i=1}^N\left(\hat{\boldsymbol{C}}_i-C_i\right)^2
\end{equation}
In addition, we propose a lidar points projection loss to supervise our rendering depth and enhance the geometry consistency:
\begin{equation}
\mathcal{L}_d=\frac{1}{\left|R_i\right|} \sum_{i \in R_i}\left(\hat{\boldsymbol{D}_i}-\Pi(P,\{R,t\},K)\right)^2
\end{equation}
where $\Pi()$ is the projection from 3D point $P_i$ to the depth of 2D pixel. $R_i$ is the set of rays with a valid depth projection value. Inspired by \cite{dnerf}, we also use ray distribution loss to minimize the KL divergence between the
rendered ray distribution $h(t)$ and the noisy depth distribution:
\begin{equation}
    \mathcal{L}_{\sigma}=\sum_{i}\log{h_{i}}\mathrm{exp} (-\frac{(t_i-\hat{\boldsymbol{D}_i})^2}{2\hat{\sigma}_i^2})\delta_i
\end{equation}
where $\hat{\sigma}_i^2$ is the variance of depth $D_i$.

\noindent\textbf{Far-Field Modeling (Sky).}
Outdoor scenes contain sky regions where rays never intersect any opaque surfaces, and thus, the NeRF model gets a weak supervisory signal in those regions.

To modulate which rays utilize the environment map, we run the segment anything model (SAM) for each image to detect pixels of the sky: $S_i=S(I_i)$,
where $S_i(r)=1$ if the ray $r$ goes through a sky pixel in image $i$. We then design a far-field loss to
encourage at all point samples along rays through sky pixels to have zero density:
\begin{equation}
     \mathcal{L}_{\mathrm{seg}}=\mathbb{E}_{\mathbf{r} \sim I_i}\left[\mathcal{S}_i(\mathbf{r}) \int_{t_n}^{t_f} (T\cdot\alpha)^2 d t\right]
\end{equation}
At each step, the gradient at each trainable node in $L$ and $F$ intersecting with the rays is computed and backpropagated.

\section{Implementation}\label{sec:implementation}
\subsection{Dataset and Metrics}
We evaluate ProSGNeRF on KITTI tracking datasets \cite{kitti} for the real-world urban scene. The KITTI dataset \cite{kitti} is captured using a stereo camera. Each sequence consists of more than 100 time steps and images of size $1242 \times 375$,
each from two camera perspectives and more than 12 dynamic objects from different classes. 
We strictly follow the experimental setting of NSG and SUDS to ensure fairness. 
  In addition to the experiments on the KITTI dataset, we further conducted evaluations on the synthetic vKITTI2 dataset~\cite{vkitti} and the larger real-world Waymo dataset~\cite{waymo} and nuScenes~\cite{nuScenes} datasets, thereby demonstrating the outstanding performance of our model across diverse scenarios and ensuring a comprehensive evaluation.
 We report the standard view synthesis metrics
PSNR, SSIM, and LPIPS scores for all evaluations.

\noindent{}{\bf Dataset Generation}
  We construct a dataset with three widely used autonomous driving datasets including KITTI Tracking, VKITTI2, Waymo Open Dataset  using the pre-trained SAM model and a 3D object detection model, which contains camera intrinsics and extrinsics, RGB images, as well as 3D bounding boxes and 2D masks of objects. We use SAM to acquire precise 2D semantic masks without the requirement for human annotation. Leveraging the promptable segmentation method, SAM predicts masks for both dynamic objects and scenarios in far-field situations. This approach enables the construction of a progressive neural scene graph, integrating instance masks from SAM and 3D bounding boxes derived from machine predictions.

  KITTI Tracking~\cite{kitti} contains 21 training sequences with 8,008 frames, corresponding to 16,016 stereo images at a resolution of $1242 \times 375$. It provides labeled dynamic classes such as cars and pedestrians, together with 3D tracking boxes or detector boxes, while 2D masks for dynamic objects are generated using SAM2 in our pipeline. VKITTI2~\cite{vkitti} consists of 5 scenes with 10 variants, including 21,260 stereo pairs and 42,520 stereo images. It provides vehicle instance annotations, instance segmentation masks, and bounding boxes, and is mainly used for evaluation. Waymo Open Dataset~\cite{waymo} is a large-scale real-world dataset with 2,030 segments, 390,000 frames, and 1,950,000 camera images. It includes around 31K unique object instances, 1.2M object frames, panoptic masks for 100K camera images, and 12.6M 3D bounding boxes. In our experiments, Waymo is used for both evaluation and object-prior construction, where SAM-generated masks are used for the selected objects.


\subsection{Parameters}
ProSGNeRF is evaluated on a desktop PC with NVIDIA A100 GPU. For experimental settings. Each decoder uses the five fully-connected residual blocks.  The bounding box dimensions are scaled to include the shadows for better reconstruction.

\noindent\textbf{Loss weight} \quad
In refining local scene graphs, we utilize a color loss mechanism to enhance the representation of background and dynamic object nodes. Introducing a novel lidar points projection loss bolsters geometric consistency by supervising rendering depth. Drawing inspiration from \cite{dnerf}, our approach incorporates a ray distribution loss to minimize divergence between the rendered ray distribution $h(t)$ and observed noisy depth distribution.Moreover, we introduce a far-field loss function to ensure zero density for point samples along rays passing through sky pixels. The depth loss and KL divergence loss are pivotal in maintaining geometric consistency, ultimately enhancing scene accuracy and geometry portrayal. However, excessive emphasis on these losses can negatively impact image metrics. Hence, controlling their weights is crucial for optimal performance. 

In our experiments, We set $\lambda_{c}= 1$ , $\lambda_{d}= 0.005$ ,$\lambda_{\sigma}= 0.005$, $\lambda_{seg}=0.001$. These adjustments strike a balance between geometric accuracy and image fidelity, refining scene representation while minimizing adverse effects on image-related metrics.

\noindent \textbf{Baseline}  We compare our method against state-of-art implicit neural rendering methods: the original NeRF implementation~\cite{NeRF},
NeRF+Time, NSG~\cite{nsg}, SUDS~\cite{suds}, S-NeRF~\cite{snerf}, StreetSurf~\cite{streetsurf}, MARS~\cite{mars}, EmerNeRF~\cite{emernerf}, and 3D Gaussian Splatting~\cite{3dgs},  StreetGaussian~\cite{streetgaussian}, Driving Gaussian~\cite{drivinggaussian}, PVG~\cite{pvg}, DeSiRe-GS~\cite{desiregs}.

\section{Experiments}
In this section, we validate the proposed ProSGNeRF method on the existing
autonomous driving datasets KITTI \cite{kitti}, VKITTI~\cite{vkitti}, Waymo \cite{waymo} and nuScenes datasets~\cite{nuScenes}. 
We validate that our method outperforms existing dynamic representation methods in image reconstruction, novel view synthesis, and dynamic object manipulation.

\begin{figure*}[t]
  \centering
   \includegraphics[width=\linewidth]{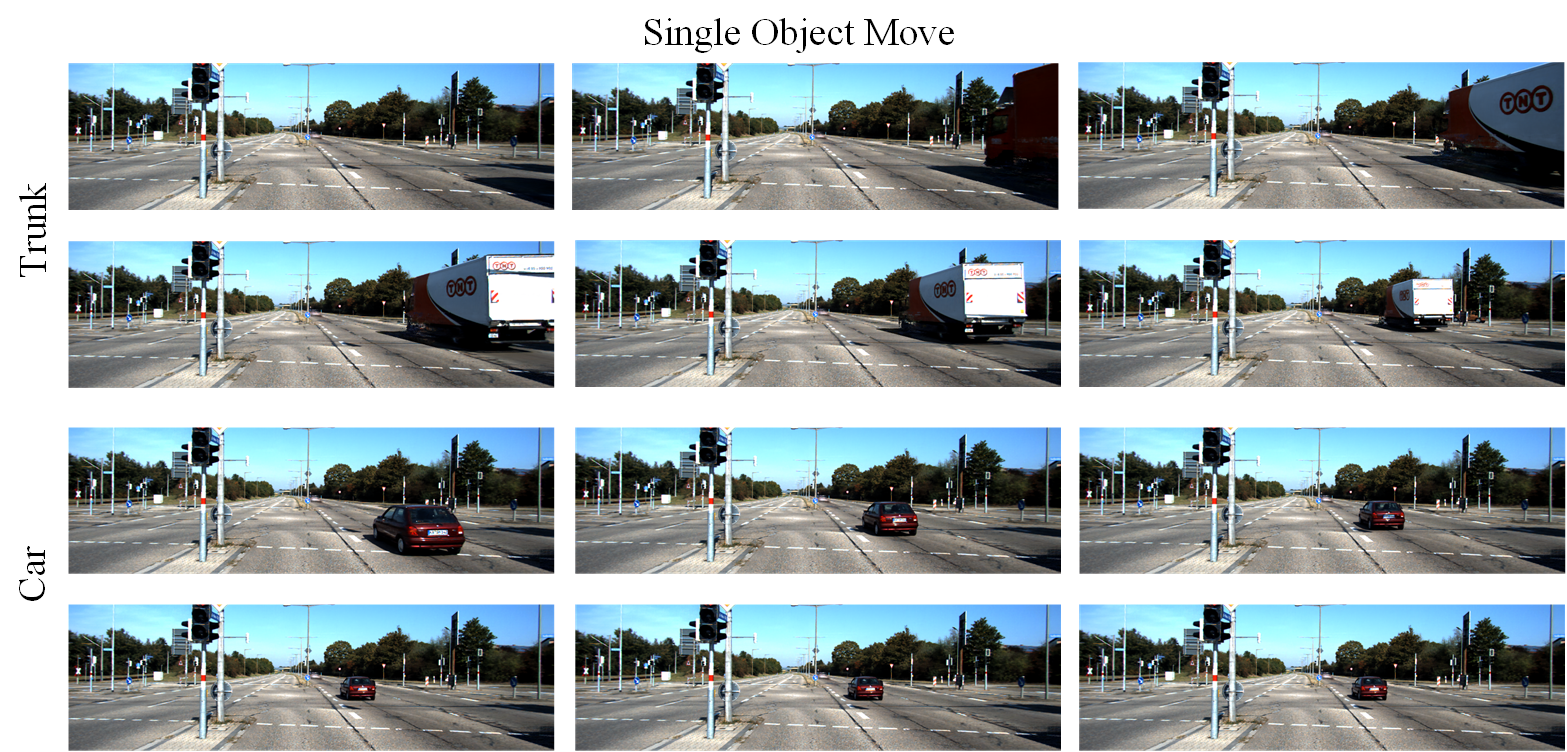}

    \caption{We illustrate the scene editing results for the single object movement setting. As shown in the figure, our method enables flexible insertion of different types of vehicles into the scene, including trucks and cars.}

   \label{fig:manipulation2}
\end{figure*}

\subsection{Experimental Results}
\noindent\textbf{Image Reconstruction and Novel View Synthesis.} We conduct image reconstruction and novel view synthesis experiments in urban dynamic scenes. Table \ref{tab:kitti} reports the qualitative results of image reconstruction and novel view synthesis on the KITTI dataset \cite{kitti}. To comprehensively evaluate performance under varying data availability, we adopt multiple train and test splits, including using the full dataset, holding out every fourth time step, holding out every other time step, and training with only one out of every four time steps. Quantitative evaluations are performed using PSNR, SSIM, and LPIPS metrics. Across all experimental settings, the proposed method consistently outperforms all baseline approaches on all evaluation metrics. Compared with NeRF based methods, our approach achieves substantially improved performance in both background reconstruction and dynamic object reconstruction. This improvement can be attributed to the progressive scene graph design, which enables large-scale ego-motion scene modeling, as well as the incorporation of multiresolution hash encoding in the background network, allowing more accurate representation of fine scale geometric and texture details. In comparison with 3DGS based methods, our approach demonstrates superior performance in dynamic object reconstruction and exhibits stronger generalization capability. Specifically, our method is able to effectively model a wide range of dynamic object categories across different scenes, highlighting its robustness and adaptability in complex dynamic environments. In particular, as the number of available training views decreases, our method exhibits a more significant advantage in novel view synthesis performance. 

In addition to KITTI, we also present image reconstruction and novel view synthesis results on the VKITTI~\cite{vkitti}, Waymo~\cite{waymo}, and nuScenes~\cite{nuScenes} datasets in Tab.~\ref{tab:kitti}, Tab.~\ref{tab:vkitti}, Tab.~\ref{tab:waymo}, and Tab.~\ref{tab:nuscenes}. The experimental results across these diverse urban driving benchmarks clearly demonstrate the effectiveness and strong generalization capability of the proposed method, showing its ability to accurately model both static background structures and dynamic objects in complex real world scenes.
The qualitative results are shown in Fig. \ref{fig: kitti} and \ref{fig:kitti0}. Figure \ref{fig: kitti} depicts the appearance of the dynamic scene at different time instances from left to right (time step: 66, 80, 102, 234).
The vanilla NeRF implicit rendering fails to reconstruct dynamic scenes accurately. It can only represent static scenes, which leads to nearly identically rendered frames for all the sequences. The NSG algorithm provides a relatively accurate representation of dynamic objects but cannot handle scenes with large-scale camera ego-motion. This limitation arises from the fixed root of their scene graph representation, which cannot move with the camera. In the last column of the figure, their reconstruction results appear highly blurred when the camera moves. SUDS exhibits poor representation capability for dynamic objects; their objects appear blurry in Fig.~\ref{fig: kitti}. Methods based on 3DGS achieve relatively high accuracy in background modeling. However, they exhibit limited performance in dynamic object reconstruction, suffer from poor generalization capability, and rely heavily on high precision point cloud initialization such as LiDAR measurements. In contrast, our method leverages prior knowledge encoded in a foundation model to guide scene understanding, enabling effective extraction of object appearance and shape features. This design allows the proposed approach to robustly handle the sparse observation problem caused by high speed object motion, while simultaneously maintaining strong generalization performance across different scenes and object categories.
Especially as the number of training views decreases, all the existing methods generate ghosting artifacts near areas of movement.  We also present object PSNR and LPIPS curves as the training view falls in Fig.~\ref{fig:manipulation3}. ProSGNeRF achieves state of the art performance in dynamic object reconstruction and novel view synthesis under large scale camera ego motion, which strongly validates the effectiveness of the proposed frequency modulated foundation model design.

\begin{table*}
\centering
\caption{Ablation study of our design choices on  KITTI datasets \cite{kitti} for image reconstruction and 50\% for novel view synthesis. The metrics are PSNR, SSIM, LPIPS.}
\scalebox{0.85}{
\setlength{\tabcolsep}{0.95mm}{
\begin{tabular}{lllllll}
\toprule
\multirow{2}{*}{Methods}                      & \multicolumn{3}{c}{Image Reconstruction}               & \multicolumn{3}{c}{Novel View Synthesis-50\%}          \\  
                                              & PSNR $\uparrow$ & SSIM $\uparrow$ & LPIPS $\downarrow$ & PSNR $\uparrow$ & SSIM $\uparrow$ & LPIPS $\downarrow$ \\ \hline
w/o progressive scene graph                   & 26.34           & 0.849           & 0.079              & 23.98           & 0.795           & 0.137              \\
w/o overlapping between local scene graph     & 28.79           & 0.862           & 0.065              & 24.98           & 0.834           & 0.103              \\ \hline
with only MLP                                 & 28.75           & 0.865           & 0.065              & 24.81           & 0.832           & 0.104              \\
with only Dino v2                        & 29.41           & 0.881           & 0.060     & 25.98           & 0.865           & 0.088              \\
with frequency modulation and MLP             & 27.18           & 0.860           & 0.068              & 24.72           & 0.831           & 0.105              \\
w/o object pose input                         & 27.98           & 0.864           & 0.071              & 24.75           & 0.822           & 0.103              \\
with shared encoders for shape and appearance & 27.15           & 0.867           & 0.073              & 24.28           & 0.816           & 0.113              \\
using the shared network for car and truck    & 27.81           & 0.877           & 0.071              & 24.56           & 0.839           & 0.101              \\ \midrule
w/o depth loss                                & 29.25           & 0.885           & 0.063              & 25.19           & 0.865           & 0.094              \\
w/o KL divergence loss                        & 29.28           & 0.891           & 0.069              & 25.45           & 0.860           & 0.096              \\
w/o far-field loss                            & \textbf{29.37}  & 0.890           & 0.067              & 25.28           & 0.862           & 0.095              \\ \hline
without planes sampling                       & 29.68           & 0.896           & 0.065              & 25.89           & 0.864           & 0.091              \\
without ray-box sampling                      & 29.72           & 0.897           & 0.063              & 25.82           & 0.865           & 0.090              \\ \hline
Full Model                                 & \textbf{30.31} & \textbf{0.931} & \textbf{0.057}     & \textbf{26.55}  & \textbf{0.878}  & \textbf{0.086}     \\ \bottomrule
\end{tabular}}}

\label{tab:ablation}
\end{table*}

\noindent\textbf{Scene Decomposition and Object Manipulation.}
Compared to other methods, our approach excels significantly in its ability for scene decomposition and object manipulation. The structure of the proposed method can naturally decompose scenes into dynamic and static scene components. We present the rendering of isolated nodes of our progressive scene graph in Fig. \ref{fig: manipulation} and Fig to present the scene editing ability. In Fig. \ref{fig: manipulation}, we compared the scene decomposition ability of our method and other methods. We can see that our method learns a clean decomposition between the objects and the background. 

We then generate previously unseen frames of novel object arrangements, temporal variations. Our progressive neural scene graph architecture allows dynamic urban scene editing, which can be manipulated at its edges and nodes. To demonstrate
the flexibility of the proposed scene representation, we rotate the pose of dynamic objects. We present the view synthesis results in Fig. \ref{fig: manipulation} and Fig.~\ref{fig:manipulation2}.
Note that we have not observed the rotation of dynamic objects during training.

In addition to object manipulation and node removal of the dynamic objects, our method allows for constructing
completely novel scene graphs and novel view synthesis.
Fig. \ref{fig: manipulation} shows view synthesis results that we randomly insert new object nodes and new object positions with new transformations. We constrain samples to the joint set of all observed road trajectories. Note that all these translations and manipulations have not been observed during training. Figure \ref{fig:manipulation2} illustrates the scene editing results for the single object movement setting. As shown in the figure, our method enables flexible insertion of different types of vehicles into the scene, including trucks and cars. These results demonstrate the strong object level editability of the proposed approach and highlight the advantages of the scene graph structure in supporting controllable and semantically consistent scene manipulation.

\noindent \textbf{Computational requirements.}
We further report the computational requirements of ProSGNeRF to clarify its practical efficiency. 
All experiments are conducted on a single NVIDIA A100 GPU. 
For a representative KITTI sequence, the training time is approximately 10 hours, the peak GPU memory consumption is 14.5 GB, and the rendering speed is 3 FPS at the original image resolution.

\subsection{Ablation Study}

In this section, we provide more detailed ablation studies of our method. The detailed experimental results is shown in Tab. \ref{tab:ablation}. 
The table presents an ablation study showing the impact of different design choices on the performance of a scene representation model. It evaluates image reconstruction and novel view synthesis on the KITTI dataset across three metrics: PSNR, SSIM, and LPIPS. 

Several variants of the model are compared by incrementally removing or changing components. This includes ablations of the progressive scene graph, loss functions, encoder sharing schemes, input modalities, and sampling strategies. The bottom row shows results for the full model with all components enabled.

The full model with all features performs the best across most evaluation metrics. The most impactful design choices are the progressive scene graph, separate encoders for shape and appearance, and ray-based sampling. Removing or changing these degrades performance, demonstrating their importance. Other ablations produce smaller differences, suggesting less critical roles.
Moreover, we present the ablation study on progressive scene representation with pre-dividing in Tab. \ref{tab:ablation1}.

Overall, the ablation quantitatively validates design decisions through comparative analysis. It highlights key model architectures and training techniques that boost reconstruction fidelity and view synthesis photorealism. The table effectively summarizes component importance across multiple metrics.

\begin{table}[h]
\centering
\caption{Ablation study on  KITTI \cite{kitti} and Waymo \cite{waymo} datasets. We demonstrate the effectiveness of progressive scene representation and online scaling.}
\setlength{\tabcolsep}{0.7mm}{
\begin{tabular}{lcccccc}
\hline
\multicolumn{1}{c}{\multirow{2}{*}{Methods}} & \multicolumn{3}{c}{KITTI \cite{kitti}}                                                         & \multicolumn{3}{c}{Waymo \cite{waymo}}                                                         \\ 
\multicolumn{1}{c}{}                         & PSNR                      & SSIM                      & LPIPS                     & PSNR                      & SSIM                      & LPIPS                     \\ \hline
pre-divide                                  & \multicolumn{1}{l}{24.89} & \multicolumn{1}{l}{0.827} & \multicolumn{1}{l}{0.118} & \multicolumn{1}{l}{31.28} & \multicolumn{1}{l}{0.908} & \multicolumn{1}{l}{0.112} \\
w/o scaling                                  & 25.29                     & 0.837                     & 0.108                     & 32.85                     & 0.915                     & 0.103                     \\
Ours                                         & \textbf{25.95}            & \textbf{0.856}            & \textbf{0.095}            & \textbf{33.80}            & \textbf{0.931}            & \textbf{0.092}            \\ \hline
\end{tabular}}

\label{tab:ablation1}

\end{table}

\noindent \textbf{Progressive scene graph.}
We use the original neural scene graph instead of our progressive scene graph, which leads to a remarkable decrease in the image accuracy metrics in dynamic urban scenes with large-scale camera ego-motion.

\noindent \textbf{Role of DINOv2 and frequency modulation.}
The DINOv2 encoder and the frequency modulation module play different roles in our dynamic object branch. 
DINOv2 provides the main high-level shape and appearance prior for sparse-view dynamic objects, while frequency modulation serves as a progressive regularization strategy for the object-conditioned input features. 
As shown in Table~\ref{tab:ablation}, using only DINOv2 already provides strong performance, indicating the importance of foundation-model object priors. 
Adding frequency modulation further improves the full model by stabilizing early optimization and gradually introducing high-frequency details. 
In contrast, applying frequency modulation to an MLP-only object branch does not provide comparable performance, because frequency scheduling alone cannot replace the semantic and geometric priors provided by DINOv2.

\noindent \textbf{Loss.} In the third and fourth experiments, we remove the depth loss and KL divergence loss from our network and compare it to the full model. The comparison indicates that the depth loss and KL divergence loss provide the geometry consistency of the scene representation, improving the scene's accuracy and geometry representation.  
The full model achieves the best image reconstruction and novel view synthesis performance.

\noindent \textbf{Window size.} In addition to our bound-based automatic allocation strategy, we evaluate fixed temporal window lengths of 30, 50, and 100 frames. 
The results show that too small windows lead to insufficient local observations and weaker cross-frame consistency, while too large windows increase memory consumption and make local optimization more difficult under large camera ego-motion. 
Our automatic bound-based allocation achieves the best trade-off between reconstruction quality and computational cost.

\begin{table}[t]
\centering
\caption{Ablation study on local window size on the KITTI sequence.}
\label{tab:window_size_ablation}
\setlength{\tabcolsep}{0.7mm}{
\begin{tabular}{lcccc}
\toprule
Window strategy & PSNR$\uparrow$ & SSIM$\uparrow$ & LPIPS$\downarrow$ & Peak memory \\
\midrule
Fixed, 30 frames  & 27.29 & 0.905 & 0.065 & 13.7 GB \\
Fixed, 50 frames  & 28.55 & 0.926 & 0.061 & 15.5 GB \\
Fixed, 100 frames & 26.95 & 0.897 & 0.073 & 18.6 GB \\
Automatic bound & 30.31 & 0.931 & 0.057 & 14.5 GB \\
\bottomrule
\end{tabular}}
\end{table}

\subsection{Limitations and Future work}
Although ProSGNeRF achieves strong performance in large-scale dynamic urban scenes, it still has several limitations. 
The current dynamic object branch mainly assumes rigid or approximately rigid moving objects, such as cars, vans, and trucks. 
Highly deformable or articulated objects, such as pedestrians and cyclists, are more challenging because their motion cannot be fully represented by a single rigid object-centric transformation. 
When reliable masks and 3D boxes are available, our method can still reconstruct their approximate appearance and motion, but the reconstruction quality is generally lower than that of rigid vehicles.

The quality of the progressive scene graph depends on the preprocessing stage, including 2D segmentation masks, 3D object boxes, and object tracking. 
Failure cases in segmentation or detection may introduce incorrect object nodes or incomplete object regions. 
Tracking failures may also break the identity consistency of dynamic objects across local scene graphs. 
For long-term occlusion or object re-entry, if the tracker fails to preserve the object identity, the re-entered object is initialized as a new dynamic node. 
Although the class-shared object decoder and foundation-model prior can still help reconstruction, strict long-term identity consistency is not guaranteed.

\section{Conclusion}
We present ProSGNeRF, a progressive dynamic neural scene graph framework for large-scale urban scene reconstruction with multiple dynamic objects. By dynamically allocating local scene graphs, decoupling static background, far-field regions, and object-centric dynamic nodes, and introducing a frequency-modulated foundation-model object branch, our method improves scalable reconstruction and novel view synthesis in challenging urban driving scenes. For dynamic objects, our frequency auto-encoder network efficiently encodes shape and appearance, addressing sparse view observations through frequency modulation. Extensive experiments on both simulated and real data achieve photo-realistic quality, showcasing the effectiveness of our approach.
We believe that this work opens up the field of neural rendering for city-scale dynamic scene reconstruction. 

 \noindent\textbf{Acknowledgements} This work is supported by the National Natural Science Foundation of China under Grant 62573287 and the Science and Technology Commission of Shanghai Municipality under Grant 20DZ2220400. (Corresponding authors: Weidong Chen. wdchen@sjtu.edu.cn)
 *The first two authors contribute equal to this paper.

\noindent\textbf{Availability of data and materials} All the datasets used in the paper will be publicly available.

\bibliography{main}
\end{document}